\documentclass[10pt,twocolumn,letterpaper]{article}

\usepackage[pagenumbers]{cvpr} 

\definecolor{cvprblue}{rgb}{0.21,0.49,0.74}

\usepackage{multirow}
\usepackage{pifont}
\usepackage[table, dvipsnames]{xcolor}
\definecolor{tblue}{HTML}{F0F8FF}
\definecolor{alogryellow}{HTML}{a07018}
\usepackage{tabularx}
\usepackage{pifont}
\usepackage{amssymb, mathtools}
\usepackage{booktabs}
\usepackage{multirow}
\usepackage{tablefootnote}
\usepackage{lineno}
\usepackage{makecell}
\definecolor{downred}{HTML}{F08080}
\definecolor{upgreen}{HTML}{3CB371}
\usepackage{algorithm}
\usepackage{algorithmic}
\newcommand{\algcomment}[1]{\textcolor{alogryellow}{/* #1 */}}
\definecolor{iccvblue}{rgb}{0.21,0.49,0.74}
\usepackage[pagebackref,breaklinks,colorlinks,allcolors=iccvblue]{hyperref}
\title{When Fish Look Alike: \\Tracking Identities with Dual-branch Elasticity}

\author{Vran Lee\textsuperscript{1}\hspace{0.2cm}
	Xin Liu\textsuperscript{2}\hspace{0.2cm}
	Yijie Wei\textsuperscript{1}\hspace{0.2cm}
	Yeqiang Liu\textsuperscript{1}\hspace{0.2cm}
	Hwa Liang Leo\textsuperscript{3}\hspace{0.2cm}
	Zhenbo Li\textsuperscript{1}\thanks{Corresponding author.}
	\vspace{1.5mm}\\
	\textsuperscript{\rm 1}China Agricultural University\hspace{0.5cm}
    \textsuperscript{\rm 2}Beijing Normal University\hspace{0.5cm}
	\textsuperscript{\rm 3}National University of Singapore
	\vspace{1mm}\\
	{\tt\small vranlee86@gmail.com, lizb@cau.edu.cn}\\
}

\begin{document}
\maketitle
\begin{abstract}
Tracking dense, homogeneous targets like schooling fish remains a major challenge for multiple object tracking due to extreme inter-individual homogeneity, severe physical clustering, and rapid non-rigid deformations. While heavy-backbone separated detection and embedding trackers like SU-T push accuracy boundaries using complex Re-Identification networks, their computational overhead prohibits edge deployment. Furthermore, these modules often fail when appearance features degrade under severe occlusions. To overcome this, we propose \textbf{T}racking \textbf{I}dentities with \textbf{D}ual-branch \textbf{E}lasticity (\textbf{TIDE}). Bypassing expensive appearance cues, TIDE utilizes the Adaptive Geometric Correspondence IoU, an association mechanism leveraging spatial and structural consistency to robustly handle complex morphological variations. Crucially, TIDE introduces system-level deployment elasticity, decoupling the algorithmic pipeline from strict hardware constraints. Evaluations on the MFT-Edge benchmark demonstrate that our Lightweight L-branch achieves a competitive HOTA of 28.43 using merely 20.47G FLOPs. This represents a 38.7-fold computational reduction compared to upper bounds like SU-T, directly facilitating real-time edge deployment. Concurrently, our Scalable S-branch establishes a 29.98 HOTA, successfully bridging the gap between high-precision cloud analysis and efficient edge tracking. The dataset and codes are released at \textit{\href{https://vranlee.github.io/TIDE/}{https://vranlee.github.io/TIDE/}}.
\end{abstract}    
\section{Introduction}

\textit{To keep that date in a sea of identical silhouettes, exhaustive scrutiny becomes a needless burden. The true match is often recognized with a single \textbf{look}, unfolding a connection that is \textbf{swift and effortless}.}
\begin{flushright}
	\textit{(Preface)}
\end{flushright}

\begin{figure}[t]
	\centering
	\includegraphics[width=\linewidth]{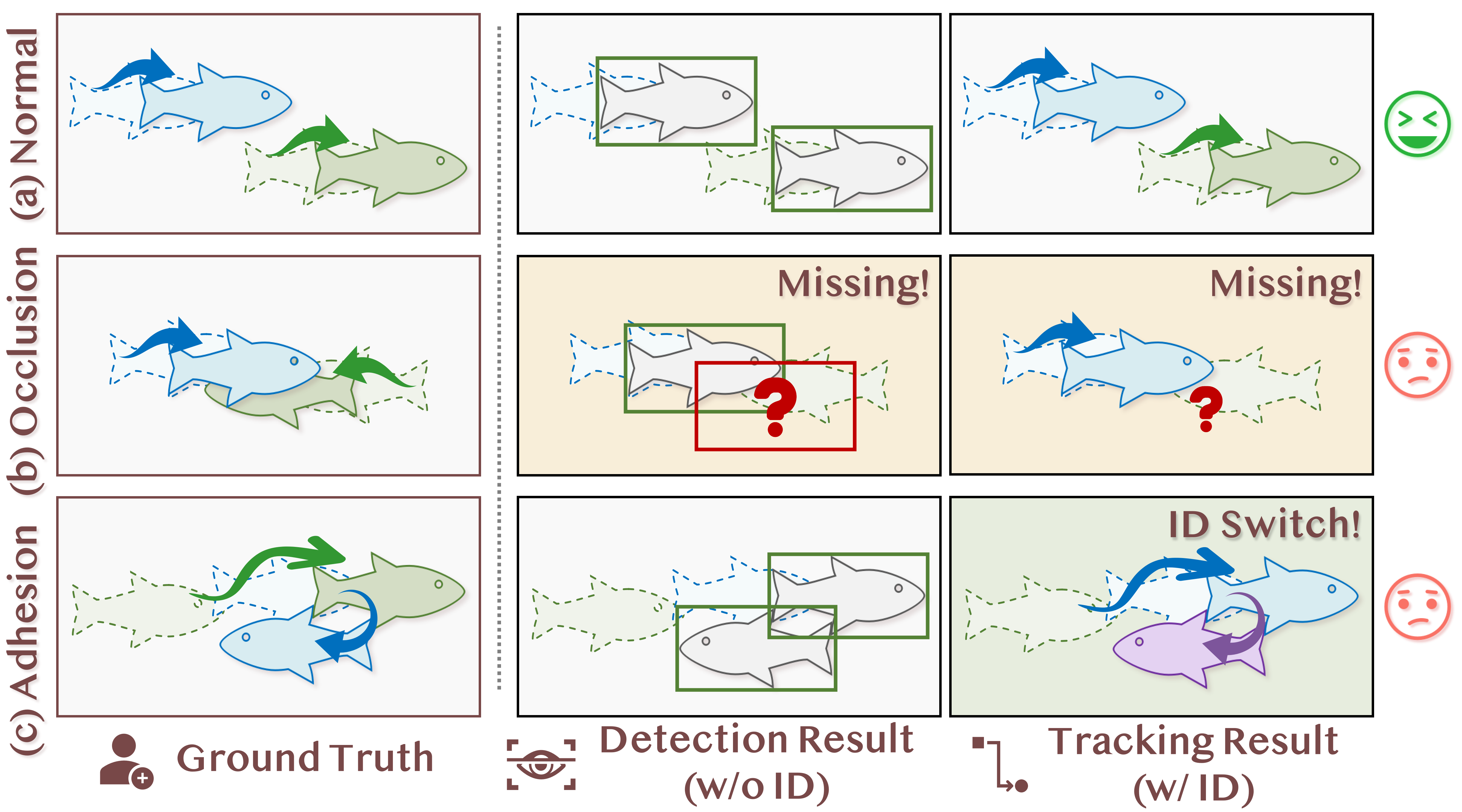}
	\caption{Key challenges in multiple fish tracking. The figure compares (a) a successful tracking scenario with common failure modes: (b) a missed detection, where a fish is not localized, and (c) an identity switch, where two tracks are incorrectly swapped. Best viewed in color.}
	\label{fig:challenges}
\end{figure}

\begin{figure}[t]
	\centering
	\includegraphics[width=\linewidth]{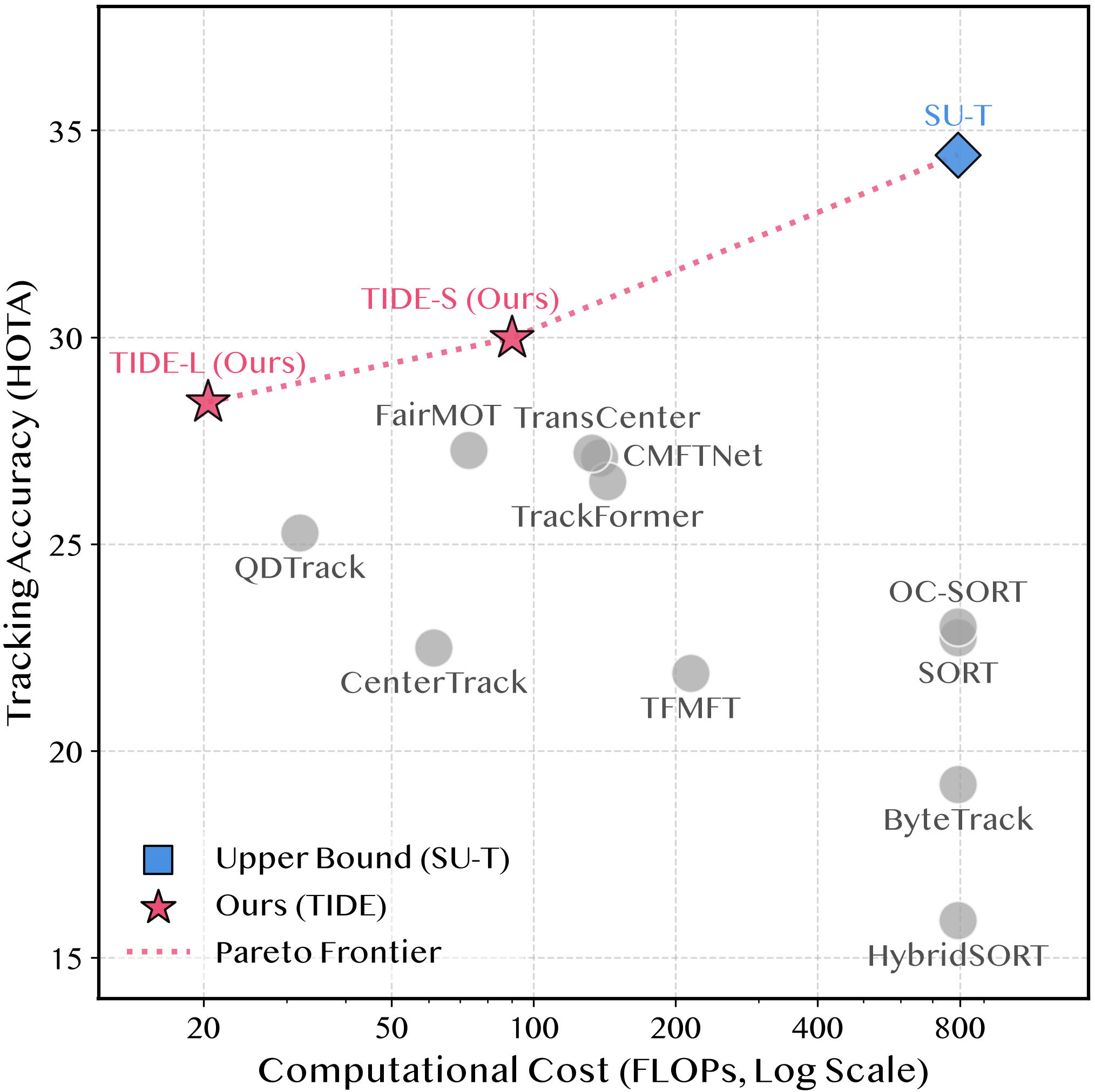}
	\caption{Comprehensive comparison of the proposed TIDE against specialized MFT and general MOT methods.}
	\label{fig:SOTA}
\end{figure}

Multiple Object Tracking (MOT) has achieved remarkable success in rigid and distinct scenarios, such as autonomous driving~\cite{zhang2021fairmot, seidenschwarz2023simple}. However, tracking dense, homogeneous, and highly deformable targets, including schooling fish, remains a significant challenge for modern visual systems. These scenarios are characterized by minimal inter-individual variance, severe physical adhesion, and rapid non-rigid deformations (e.g., C-shaped bursts)~\cite{li2024tfmft, li2022cmftnet}, as illustrated in Fig.~\ref{fig:challenges}. Recognizing these complexities, recent work~\cite{li2026trackers} established MFT25, a large-scale universal benchmark and baseline framework, SU-T, which advances the accuracy boundaries of underwater Multiple Fish Tracking (MFT).

While SU-T successfully improves performance limits on high-end hardware, a critical dilemma persists for real-world industrial deployment: substantial computational overhead. The dominant Separated Detection and Embedding (SDE) paradigms utilized in standard benchmarks rely heavily on complex Re-Identification (Re-ID) modules to distinguish visually similar targets~\cite{zhang2022bytetrack, zhao2023object}. In dense scenarios, this computational burden scales linearly with object counts, rendering them impractical to deploy on resource-constrained edge devices, such as underwater IoT cameras. Conversely, Joint Detection and Embedding (JDE) architectures decouple computation from crowd density but often suffer from unstable associations when appearance features degrade during severe occlusions. Furthermore, attempts to mitigate ID switches via intricate morphological IoU variants or non-linear motion models incur significant mathematical overhead.

To bridge the gap between academic accuracy limits and real-world edge deployment, particularly for management systems in active fishing grounds, we propose \textbf{TIDE} (\textbf{T}racking \textbf{I}dentities with \textbf{D}ual-branch \textbf{E}lasticity) as an applied, highly efficient sister framework to the SU-T baseline. Addressing the challenges of extreme inter-individual homogeneity and severe physical adhesion, TIDE bypasses computationally intensive Re-ID networks. Instead, we introduce the Adaptive Geometric Correspondence IoU (AGCIoU), an association mechanism that leverages robust centroid proximity and structural consistency to maintain identities through severe occlusions with minimal computational burden.

Additionally, to accommodate varying hardware constraints without introducing the complex control flows and memory overheads typical of traditional elastic networks, TIDE utilizes a streamlined, modular architecture. By unifying a downstream multi-task head with the AGCIoU associator, the framework provides a Lightweight branch (L-branch) tailored for extreme edge efficiency, alongside a Scalable branch (S-branch) for higher-precision analysis. Finally, we evaluate TIDE on two complementary datasets. We introduce MFT-Edge, a compact stress-test dataset targeting extreme aquatic disturbances, such as dynamic shadows and severe mirror reflections. Crucially, to ensure cross-domain robustness and algorithmic scalability, we also demonstrate the framework's strong generalization capabilities on the large-scale MFT25 benchmark.

Our main contributions are summarized as follows:
\begin{itemize}
    \item We propose TIDE, a highly efficient JDE framework that effectively resolves the computational bottlenecks of tracking dense, homogeneous targets, providing a scalable dual-branch design to accommodate diverse hardware constraints.
    \item We introduce AGCIoU, a geometric association metric that maintains robust ID consistency under severe non-rigid deformations and occlusions, completely avoiding the substantial overhead of heavy appearance models.
    \item Extensive evaluations demonstrate the framework's exceptional accuracy-efficiency balance. The lightweight TIDE-L achieves a competitive HOTA of 28.43 while reducing computational cost by 38.7-fold compared to standard heavy trackers.
\end{itemize}
\section{Related Work}
\subsection{Multiple Object Tracking in High-Density Environments}

Online MOT typically prioritizes real-time surveillance performance~\cite{cui2022joint, han2024benchmarking}. Recent advancements have enhanced tracking accuracy by integrating advanced appearance models~\cite{liu2023collaborative, li2023ovtrack} or leveraging Vision Transformers for long-range global context~\cite{hou2024salience, hu2023stdformer}. However, applying these SDE paradigms to dense, homogeneous swarms, such as schooling fish, reveals a notable limitation: their computational overhead scales linearly with crowd density, often restricting edge deployment~\cite{li2026trackers}.

Conversely, JDE architectures~\cite{wang2020towards, zhang2021fairmot, li2022cmftnet} decouple computation from density by extracting appearance features directly during the detector's forward pass. Furthermore, to address the physical adhesion and frequent occlusions inherent in dense crowds, density map regression~\cite{ren2022countingmot, sun2023indiscernible} has been utilized to extract stable spatial distributions. Building upon these concepts, our TIDE framework extends the JDE paradigm. By integrating a density-aware multi-task head within a scalable dual-branch architecture, TIDE provides flexible deployment configurations, addressing the requirements of both resource-constrained edge devices and high-capacity cloud servers.

\subsection{Trajectory Association and Geometric Metrics}
Robust trajectory association typically relies on frameworks like SORT~\cite{bewley2016simple}, which combine Kalman filtering with Hungarian matching. To mitigate ID switches among visually similar targets, modern SDE trackers often append complex Re-ID networks~\cite{zhao2023object, zhang2024magic} or rely on overlap-based IoU metrics~\cite{zhang2022bytetrack, cao2023observation}.

However, standard overlap metrics heavily depend on enclosing geometric hulls (e.g., GIoU~\cite{rezatofighi2019generalized}) or stable aspect ratio assumptions (e.g., DIoU~\cite{zheng2020distance}, CIoU~\cite{zheng2021enhancing}). These mechanisms can introduce inconsistent bounding-box penalties when targets undergo significant non-rigid deformations, such as the rapid C-shaped bending observed in swimming fish. While some domain-specific trackers address this by introducing detailed morphological contour algorithms or non-linear motion filters, they often impose substantial mathematical overhead for edge devices. To overcome this limitation, we propose AGCIoU. It substitutes complex contour calculations with essential geometric cues, namely centroid proximity and structural consistency, to maintain identities. This approach facilitates robust trajectory matching during rapid deformations with minimal additional computational cost.
\section{Methodology}

\subsection{Framework}

\begin{figure}[t]
	\centering
	\includegraphics[width=\linewidth]{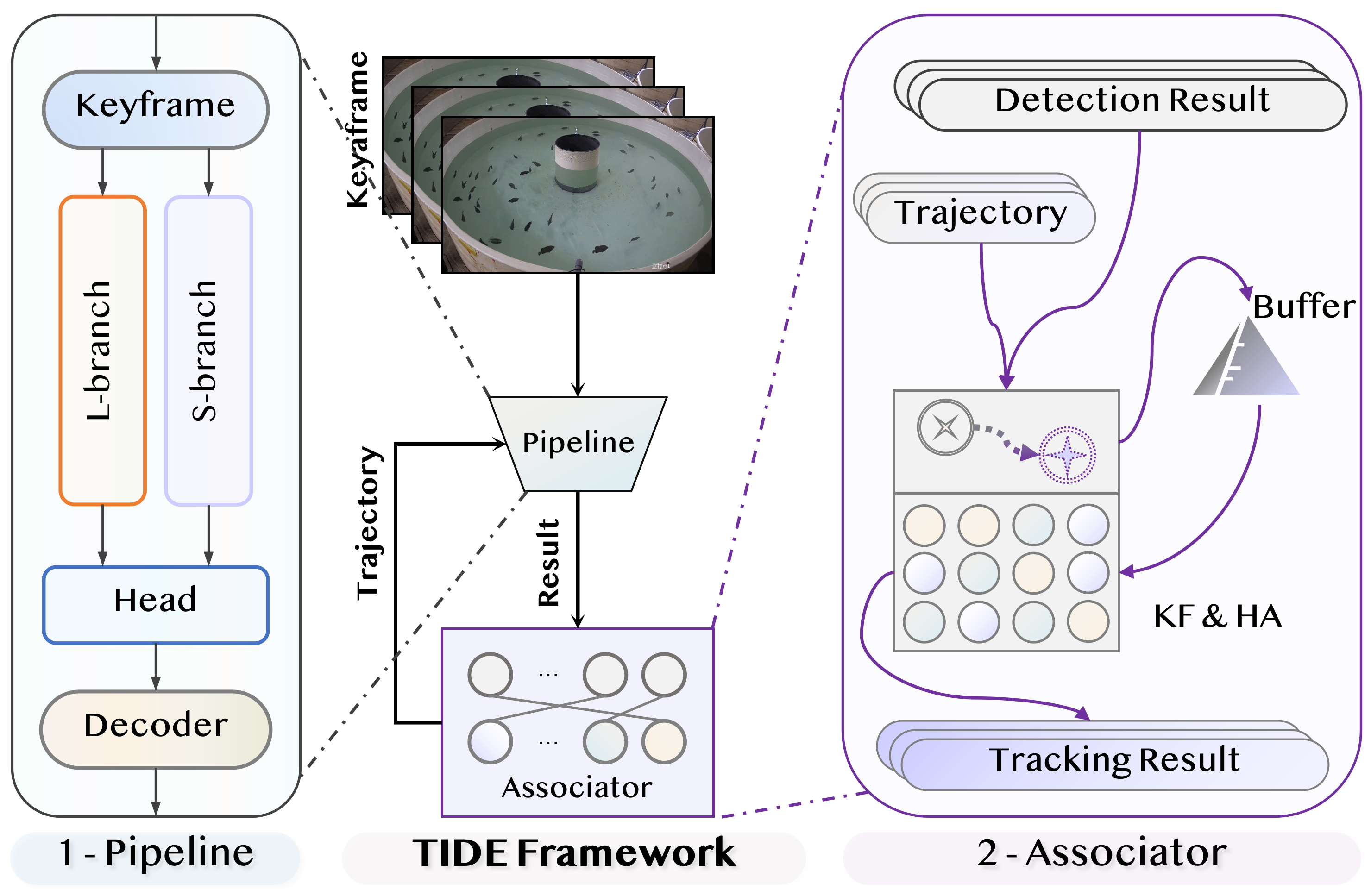}
	\caption{TIDE framework. Input frames are processed by a computationally scalable dual-branch pipeline, providing robust detections and appearance embeddings to the cascaded trajectory associator.}
	\label{fig:Framework}
\end{figure}

We propose TIDE, a highly efficient MFT framework rooted in the JDE paradigm. By extracting localized detections and appearance embeddings in a single forward pass, TIDE inherently decouples computational complexity from crowd density, effectively mitigating the bottlenecks of standard SDE approaches in dense swarms. As illustrated in Fig.~\ref{fig:Framework}, the framework consists of a scalable feature extraction pipeline and a cascaded trajectory associator. This associator integrates our minimalist AGCIoU metric, facilitating robust identity association during significant non-rigid deformations while minimizing morphological overhead.

\subsection{Dual-branch Pipeline}
\label{sec:Pipeline}

\begin{figure}[t]
	\centering
	\includegraphics[width=\linewidth]{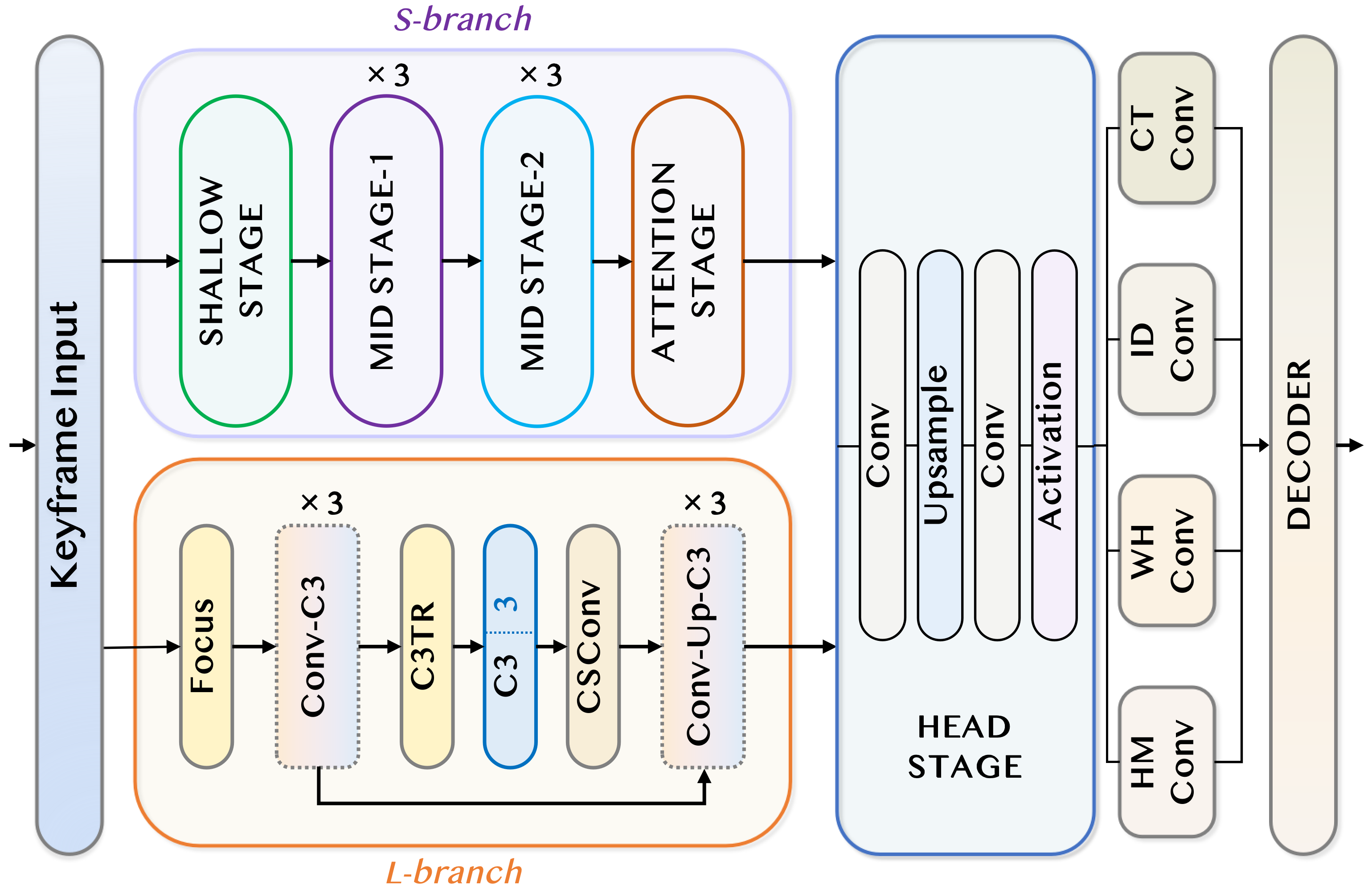}
	\caption{Overview of the dual-branch feature extraction pipeline. To adapt to varying hardware capacities, TIDE provides a Scalable S-branch (deep CNN-Transformer hybrid) for server-side precision and a Lightweight L-branch (streamlined CNN) for real-time edge deployment. Detailed layer-wise configurations are provided in the \textit{Supplementary Material}.}
	\label{fig:Pipeline}
\end{figure}

To realize system-level deployment elasticity, our feature extraction pipeline (Fig.~\ref{fig:Pipeline}) is built upon two distinct modular backbones under a unified design philosophy. Rather than operating as isolated models, both the S-branch and L-branch feed into a shared, density-aware multi-task head and utilize the identical AGCIoU associator. This modularity provides TIDE with scalable computational footprints tailored prior to deployment for specific hardware capacities, from edge nodes to cloud servers, while preserving a consistent trajectory logic.

\subsubsection{Scalable branch (S-branch)}
The S-branch is engineered to achieve high accuracy on capable servers. It employs a deep, hybrid CNN-Transformer architecture. Progressing through shallow to mid-level stages, it utilizes large $7\times7$ convolutions to significantly expand the receptive field with minimal parameter increase, effectively capturing the complete target morphology even at a distance. A final attention stage leverages Vision Transformers with positional encoding to model long-range global dependencies, which is critical for distinguishing highly homogeneous instances in occluded scenarios.

\subsubsection{Lightweight branch (L-branch)} 
The L-branch is specifically streamlined for real-time edge deployment on resource-constrained hardware (e.g., underwater IoT cameras and ROVs). It initiates with a Focus operation for information-preserving downsampling, followed by efficient spatial feature extraction using a sequence of Conv-C3 groups. To achieve robust multi-scale fusion without the substantial computational cost of traditional feature pyramid networks, we introduce the Conv-Up-C3 module, which fuses deep semantic features with shallow high-resolution maps via concatenation. Finally, tracking accuracy is refined within tight efficiency boundaries using a C3TR module (efficiently integrating local attention) and a CSConv module (reducing spatial and channel redundancy).

\subsubsection{Shared Multi-task Head}
Both branches converge into a shared head stage, utilizing standard convolutions to simultaneously output four parallel sub-tasks: heatmap generation, bounding box regression, appearance feature extraction, and density map counting. A unified decoder extracts the final instances alongside scene-level counts.

\subsection{Trajectory Associator with AGCIoU}
\label{sec:Trajectory Associator}

Traditional overlap-based metrics rely on enclosing geometric hulls or stable aspect ratio assumptions. These mechanisms can introduce inconsistent bounding-box penalties when targets undergo significant non-rigid deformations, such as rapid C-shaped bending. To address this, we introduce the AGCIoU. This minimalist metric substitutes complex hull calculations with basic geometric correspondence established through three low-overhead cues: centroid proximity, structural consistency, and adaptive weighting.

Given two bounding boxes with an intersection-over-union ($\mathrm{IoU}$), we first establish centroid proximity by calculating the normalized Euclidean center distance $\delta$:
\begin{equation}
	\footnotesize
	\delta = \frac{d_c}{s + \varepsilon}
\end{equation}
where $d_c$ is the absolute center distance, $s$ is the average scale of the two boxes, and $\varepsilon$ is a small constant. Next, we enforce structural consistency by evaluating the aspect ratio $r_i = w_i / (h_i + \varepsilon)$. We formulate an aspect ratio alignment factor $\alpha$ and an area size alignment factor $s_r$ as follows:
\begin{equation}
	\footnotesize
	\alpha = \frac{\min(r_1,r_2)}{\max(r_1,r_2)}, \quad s_r = \frac{\min(A_1,A_2)}{\max(A_1,A_2) + \varepsilon}
\end{equation}
where $A_1$ and $A_2$ denote the bounding box areas. Based on these geometric features, we define a distance penalty $\phi_d$ and a shape penalty $\phi_s$:
\begin{equation}
	\footnotesize
	\phi_d = \frac{k_d}{\delta + \varepsilon}, \quad \phi_s = k_s\,(\alpha + s_r)
\end{equation}
We then formulate AGCIoU as a piecewise function. Here, the term \textit{Adaptive} refers to its state-aware activation mechanism rather than learnable weights: it adaptively shifts the matching focus to geometric cues only when the standard IoU degrades due to severe deformations (falling below threshold $\tau$). The comprehensive mathematical derivations for these components are provided in the \textit{Supplementary Material}.

\begin{equation}
	\footnotesize
	AGCIoU = 
	\begin{cases}
		\mathrm{IoU}, & \mathrm{IoU} > \tau; \\[0.5ex]
		\mathrm{IoU} + \lambda\,(\phi_d + \phi_s), & \mathrm{IoU} \le \tau.
	\end{cases}
	\label{eq:p2piou}
\end{equation}

The trigger threshold ($\tau=0.7$) and scaling hyperparameters ($k_d=0.1, k_s=0.05, \lambda=0.5$) were empirically derived. Fixing these constants guarantees deterministic association behavior and eliminates the computational overhead of dynamic parameter generation, ensuring the mechanism remains strictly edge-friendly.

\begin{algorithm}[t]
\caption{Cascaded Trajectory Association with AGCIoU}
\label{alg:association}
\begin{algorithmic}[1]
	\REQUIRE Detections $D$, Embeddings $E$, Track Pool $P$, $\tau_{high}$, $\tau_{low}$
	\ENSURE Updated Track Pool $P$
	\item[\textbf{Notation:}] KFP = KalmanFilterPredict; CD = CosineDistance; HA = HungarianAlgorithm.
	\STATE \algcomment{\textit{\uppercase\expandafter{\romannumeral 1}. Motion Prediction}}
	\STATE $P \longleftarrow$ Current Tracks $\cup$ Lost Tracks
	\STATE $P_{pred} \longleftarrow \text{KFP}(P)$
	\STATE \algcomment{\textit{\uppercase\expandafter{\romannumeral 2}. Stage 1: Appearance-based Matching}}
	\STATE $I_{app} \longleftarrow \text{CD}(D, E, P_{pred})$
	\STATE $T_{match1}, T_{unmatch}, D_{unmatch} \longleftarrow \text{HA}(I_{app})$
	\STATE \algcomment{\textit{\uppercase\expandafter{\romannumeral 3}. Stage 2: Geometric Matching (High-Confidence)}}
	\STATE $D_{high} \longleftarrow \{d \in D_{unmatch} \mid \text{score}(d) > \tau_{high}\}$
	\STATE $I_{geo1} \longleftarrow \text{AGCIoU}(D_{high}, T_{unmatch})$
	\STATE $T_{match2}, T_{unmatch}, D_{high\_unmatch} \longleftarrow \text{HA}(I_{geo1})$
	\STATE \algcomment{\textit{\uppercase\expandafter{\romannumeral 4}. Stage 2: Geometric Matching (Low-Confidence)}}
	\STATE $D_{low} \longleftarrow \{d \in D_{unmatch} \mid \tau_{low} < \text{score}(d) \le \tau_{high}\}$
	\STATE $I_{geo2} \longleftarrow \text{AGCIoU}(D_{low}, T_{unmatch})$
	\STATE $T_{match3}, T_{unmatch}, D_{low\_unmatch} \longleftarrow \text{HA}(I_{geo2})$
	\STATE \algcomment{\textit{\uppercase\expandafter{\romannumeral 5}. Track Management}}
	\STATE Update $P$ with matched tracks $(T_{match1} \cup T_{match2} \cup T_{match3})$
	\FOR{$i \in D_{high\_unmatch}$}
	\STATE Initialize New Track$(i)$ into $P$
	\ENDFOR
	\STATE Mark remaining $T_{unmatch}$ as lost, \newline \quad remove tracks exceeding lifespan
	\RETURN $P$
\end{algorithmic}
\end{algorithm}

To systematically link detections into coherent trajectories, our associator implements a unified, cascaded two-stage recovery strategy, detailed in Algorithm~\ref{alg:association}. The first stage performs initial matching by prioritizing individuals with distinct appearance embeddings via cosine distance and Hungarian matching. The second stage handles the remaining visually ambiguous targets using our AGCIoU metric. By stratifying unmatched detections into high- and low-confidence tiers, the geometric associator recovers occluded tracks and stabilizes fragmented identities, completing a robust association loop.

\subsection{Multi-task Training}
\label{Multi-task Training}

TIDE is trained end-to-end utilizing a joint multi-task objective function that simultaneously optimizes target localization ($L_{hm}$), bounding box regression ($L_{bs+os}$), identity classification ($L_{id}$), and density map counting ($L_{ct}$). Specifically, the density map regression is supervised by penalizing predictions against the ground truth using Mean Squared Error (MSE) and the Structural Similarity Index (SSIM):
\begin{equation}
	\footnotesize
	L_{ct}=\mathrm{MSE}(\hat{D},D)+\mathrm{SSIM}(\hat{D},D)
\end{equation}
Following modern MOT paradigms~\cite{ren2022countingmot}, auxiliary constraints are applied to mutually regularize detection and counting losses to prevent redundant tracking:
\begin{equation}
	\footnotesize
	L_{\tilde{det}}=\mathrm{MSE}(D^{\prime},\hat{D}), \quad
	L_{\tilde{ct}}=\mathrm{MSE}(M^{\prime},\hat{M})
\end{equation}
where $M^{\prime}$ and $\hat{M}$ denote sliding windows of products by the detection and density map, respectively. 

To balance the gradient contributions across these diverse tasks without relying on exhaustive empirical hyperparameter tuning, we adopt a dynamic loss weighting strategy based on homoscedastic task uncertainty~\cite{kendall2018multi}. Instead of fixed linear scaling, we model the uncertainty of each task to adaptively attenuate the weights of noisier tasks. By introducing learnable log-variance parameters, the final fusion loss $L_{fuse}$ is formulated as:
\begin{equation}
	\footnotesize
	\begin{aligned}
		L_{det}^* &= L_{hm} + L_{bs+os} + L_{\tilde{det}} \\
		L_{ct}^* &= L_{ct} + L_{\tilde{ct}} \\
		L_{fuse} &= \frac{1}{2} \sum_{k \in \{det, id, ct\}} \left( e^{-s_k} L_k^* + s_k \right)
	\end{aligned}
	\label{eq:total}
\end{equation}
Here, $s_k$ represents the learnable uncertainty parameter for each task group ($det$, $id$, and $ct$). By continuously updating $s_k$ via backpropagation, the network dynamically assigns lower weights (higher uncertainty) to tasks with inherently larger gradient magnitudes or higher noise (e.g., MSE-based counting), preventing them from overwhelming the representation learning of precision-sensitive branches like appearance classification. This adaptive regularization ensures stable and optimal convergence during end-to-end training.
\begin{table*}[t]
	\centering
	\small
	\setlength{\tabcolsep}{4pt} 
	\begin{tabular*}{\linewidth}{@{\extracolsep{\fill}} l|cc|cc|ccccccc}
		\toprule
		\textbf{Methods} & \textbf{Params$\downarrow$} & \textbf{FLOPs$\downarrow$} & \textbf{HOTA$\uparrow$} & \textbf{IDF1$\uparrow$} & \textbf{IDP$\uparrow$} & \textbf{IDR$\uparrow$} & \textbf{DetRe$\uparrow$} & \textbf{DetPr$\uparrow$} & \textbf{IDs$\downarrow$} & \textbf{MOTA$\uparrow$} & \textbf{MOTP$\uparrow$} \\ \midrule
		SORT       & 99.00M & 793.21G & 22.73 & 23.91 & 29.09 & 20.29 & 44.66 & 64.03 & 2599 & 48.67 & 72.01 \\
		ByteTrack  & 99.00M & 793.21G & 19.18 & 19.37 & 26.11 & 15.40 & 35.66 & 60.46 & 2325 & 40.17 & 67.99 \\
		OC-SORT    & 99.00M & 793.21G & 22.99 & 24.14 & 29.28 & 20.54 & 44.84 & 63.92 & 2674 & 48.44 & 72.17 \\
		HybridSORT & 99.00M & 793.21G & 15.89 & 17.29 & \textbf{56.77} & 10.20 & 11.79 & 65.58 & \textbf{214} & 14.23 & 71.64 \\
		QDTrack           & 57.20M & \underline{32.02G}  & 25.27 & 24.49 & 27.74 & 21.93 & \underline{53.70} & 67.92 & 9103 & 42.81 & \underline{75.34} \\
		FairMOT           & \underline{16.55M} & 72.93G  & 27.26 & 29.68 & 36.56 & 24.98 & 46.71 & 68.36 & 2456 & 60.74 & 69.59 \\
		CMFTNet           & 45.08M & 137.77G & 27.08 & 29.93 & 36.35 & 25.43 & 47.52 & 67.93 & 2716 & \underline{61.90} & 69.47 \\
		TrackFormer       & 42.95M & 143.43G & 26.51 & 26.73 & 35.69 & 21.36 & 42.04 & \underline{70.23} & 899 & 43.42 & \textbf{76.00} \\
		CenterTrack       & 16.67M & 61.36G  & 22.49 & 23.39 & 30.90 & 18.81 & 35.11 & 57.67 & 1032 & 26.68 & 68.48 \\
		TransCenter       & 30.66M & 133.09G & 27.20 & 29.48 & 37.05 & 24.48 & 38.22 & 57.85 & 597 & 24.69 & 73.83 \\
		TFMFT             & 39.93M & 215.27G & 21.88 & 26.74 & 45.55 & 18.92 & 29.72 & \textbf{71.54} & 945 & 35.65 & 74.89 \\ \midrule
		SU-T              & 99.00M & 793.21G & \textbf{34.41} & \textbf{40.50} & 37.97 & \textbf{43.41} & \textbf{67.45} & 59.00 & 1902 & \textbf{68.52} & 71.81 \\ 
		\midrule
        \rowcolor{tblue} \textbf{TIDE-L (Ours)}   & \textbf{5.79M} & \textbf{20.47G} & 28.43 & 36.29 & \underline{49.44} & 28.67 & 37.34 & 64.41 & \underline{574} & 47.84 & 67.17 \\
		\rowcolor{tblue} \multicolumn{1}{r|}{\textit{$\Delta$ vs. SU-T}} & \textcolor{upgreen}{\textbf{\textit{-94.1\%}}} & \textcolor{upgreen}{\textbf{\textit{-97.4\%}}} & \textcolor{downred}{\textbf{\textit{-17.4\%}}} & \textcolor{downred}{\textbf{\textit{-10.4\%}}} & \textcolor{upgreen}{\textbf{\textit{+30.2\%}}} & \textcolor{downred}{\textbf{\textit{-33.9\%}}} & \textcolor{downred}{\textbf{\textit{-44.6\%}}} & \textcolor{upgreen}{\textbf{\textit{+9.2\%}}} & \textcolor{upgreen}{\textbf{\textit{-69.8\%}}} & \textcolor{downred}{\textbf{\textit{-30.2\%}}} & \textcolor{downred}{\textbf{\textit{-6.5\%}}} \\
		\rowcolor{tblue} \textbf{TIDE-S (Ours)}   & 32.59M & 90.13G & \underline{29.98} & \underline{39.01} & 47.87 & \underline{32.92} & 42.86 & 62.32 & 908 & 54.74 & 65.82 \\
		\rowcolor{tblue} \multicolumn{1}{r|}{\textit{$\Delta$ vs. SU-T}} & \textcolor{upgreen}{\textbf{\textit{-67.1\%}}} & \textcolor{upgreen}{\textbf{\textit{-88.6\%}}} & \textcolor{downred}{\textbf{\textit{-12.9\%}}} & \textcolor{downred}{\textbf{\textit{-3.7\%}}} & \textcolor{upgreen}{\textbf{\textit{+26.1\%}}} & \textcolor{downred}{\textbf{\textit{-24.2\%}}} & \textcolor{downred}{\textbf{\textit{-36.5\%}}} & \textcolor{upgreen}{\textbf{\textit{+5.6\%}}} & \textcolor{downred}{\textbf{\textit{-52.3\%}}} & \textcolor{downred}{\textbf{\textit{-20.1\%}}} & \textcolor{downred}{\textbf{\textit{-8.3\%}}} \\ \bottomrule
	\end{tabular*}
	\caption{Comparison of state-of-the-art methods. The best results are highlighted in \textbf{bold}, and the second-best results are \underline{underlined}. TIDE is shown in its Scalable (S) and Lightweight (L) configurations, with relative performance changes against the heavy baseline (SU-T) appended below. Same as follows.} 
	\label{tab:sota}
\end{table*}

\subsection{Implementation Details}
\label{sec:Implementation Details}

To validate our approach, we evaluated TIDE on two complementary datasets: the MFT-Edge stress-test dataset and the large-scale MFT25 benchmark~\cite{li2026trackers}.

MFT-Edge rigorously assesses edge architectures under extreme conditions without the evaluation overhead of massive datasets, maintaining strict environmental and annotation consistency with MFT25~\cite{li2026trackers}. Its 10 highly degraded sequences feature severe optical interferences (e.g., dynamic shadows, extreme low light, and mirror reflections) typical of recirculating aquaculture systems. This compact scale prevents lightweight models from simply memorizing training distributions, forcing reliance on robust geometric association. Conversely, we utilized MFT25 to verify cross-domain generalization across diverse aquatic environments, species, and densities.

For evaluation, we adopted the HOTA~\cite{luiten2021hota} and CLEAR~\cite{bernardin2008evaluating} metrics, prioritizing HOTA and IDF1 to balance detection accuracy and long-term identity consistency. Computational efficiency was quantified via Params and FLOPs.

All experiments were conducted on a single NVIDIA A100 GPU. Models were trained using optimal hyperparameters (e.g., $\alpha=2$ and $\lambda=0.1$ for the multi-task loss). Crucially, all baseline methods were fully adapted and fine-tuned on the respective training sets prior to evaluation to ensure strictly fair comparisons. 

\subsection{Comparison with State-of-the-Art Methods}

We benchmarked TIDE against a wide spectrum of tracking paradigms, including SDE-based methods: SORT~\cite{bewley2016simple}, ByteTrack~\cite{zhang2022bytetrack}, OC-SORT~\cite{cao2023observation}, and HybridSORT~\cite{yang2024hybrid}; JDE-style frameworks: FairMOT~\cite{zhang2021fairmot}, CMFTNet~\cite{li2022cmftnet}, and TFMFT~\cite{li2024tfmft}; as well as Transformer-based approaches: QDTrack~\cite{fischer2023qdtrack}, TrackFormer~\cite{meinhardt2022trackformer}, CenterTrack~\cite{zhou2020tracking}, and TransCenter~\cite{xu2022transcenter}. 

As detailed in Table~\ref{tab:sota}, TIDE demonstrates a superior accuracy-efficiency tradeoff. Compared to the heavy SDE upper-bound (SU-T), TIDE-L trades an expected 17.4\% drop in HOTA for massive reductions in parameters (94.1\%), FLOPs (97.4\%), and ID switches (69.8\%), directly enabling real-time inference on edge platforms like AUVs. Furthermore, other mainstream baselines struggled: HybridSORT's~\cite{yang2024hybrid} low ID switches merely reflect its severe missed detection rate, while JDE counterparts like CMFTNet~\cite{li2022cmftnet} degraded under occlusions. In contrast, TIDE excels comprehensively. The L-branch maintains a competitive HOTA (28.43) and IDF1 (36.29) at minimal cost, whereas the S-branch pushes performance to 29.98 HOTA and 39.01 IDF1, validating its deep hybrid architecture for server-side precision.

\subsection{Ablation Studies}
\subsubsection{Dual-Branch Design}
\label{Feature Extraction}

To validate our dual-branch architecture, we replaced our backbones with mainstream MOT feature extractors (e.g., DLA~\cite{yu2018deep}, ResNet-DCN~\cite{xiong2024efficient}, and ResNet-DCN-FPN~\cite{lin2017feature}) while keeping the rest of the framework constant. As shown in Table~\ref{tab:backbone}, the standard ResNet-DCN-FPN-50~\cite{lin2017feature} backbone achieves the highest HOTA score (28.41). However, our proposed Scalable S-branch demonstrates a deliberate design trade-off prioritizing long-term identity maintenance. In dense schooling environments, minimizing identity fragmentation is often prioritized over marginal bounding-box precision. By achieving the highest IDF1 (31.27) and notably the lowest number of ID switches (2131), the S-branch proves its superiority in resolving complex occlusions. Meanwhile, our Lightweight L-branch is by far the most efficient in terms of Params and FLOPs, confirming the success of its streamlined design.

\begin{table}[t]
	\centering
	\small
	\setlength{\tabcolsep}{4pt} 
	
	\begin{tabular*}{\linewidth}{@{\extracolsep{\fill}} l|cc|ccc}
		\toprule
		\textbf{Backbone} & \textbf{Param$\downarrow$} & \textbf{FLOPs$\downarrow$} & \textbf{HOTA$\uparrow$} & \textbf{IDF1$\uparrow$} & \textbf{MOTA$\uparrow$} \\ \midrule
		DLA-34            & \underline{16.70M} & \underline{79.04G}  & 27.50 & 29.55 & 58.84 \\
		R-D-34            & 23.65M & 89.13G  & 27.41 & 29.78 & 61.79 \\
		R-D-50            & 26.25M & 95.35G  & 27.26 & 29.07 & \underline{61.93} \\
		R-D-F-34          & 23.87M & 91.42G  & 27.73 & 29.94 & 61.20 \\
		R-D-F-50          & 26.80M & 101.01G & \textbf{28.41} & 29.90 & \textbf{62.08} \\
		\midrule
		\rowcolor{tblue} \textbf{TIDE-L}            & \textbf{5.79M}  & \textbf{20.47G}  & \underline{27.96} & \underline{30.04} & 61.06 \\
		\rowcolor{tblue} \textbf{TIDE-S}            & 32.59M & 90.13G  & 27.77 & \textbf{31.27} & 60.22 \\ \bottomrule
	\end{tabular*}
	
	\vspace{0.15cm}
	
	\begin{tabular*}{\linewidth}{@{\extracolsep{\fill}} l|cccccc}
		\toprule
		\textbf{Backbone} & \textbf{IDP$\uparrow$} & \textbf{IDR$\uparrow$} & \textbf{DRe$\uparrow$} & \textbf{DPr$\uparrow$} & \textbf{IDs$\downarrow$} & \textbf{MOTP$\uparrow$} \\ \midrule
		DLA-34            & \underline{37.28} & 24.48 & 45.90 & \textbf{69.90} & 2579 & \textbf{71.16} \\
		R-D-34            & 36.10 & 25.35 & 48.01 & 68.37 & 2593 & 69.80 \\
		R-D-50            & 34.94 & 24.86 & \textbf{48.93} & 68.70 & 3010 & 70.63 \\
		R-D-F-34          & 36.71 & 25.28 & 47.66 & 69.22 & \underline{2555} & 70.46 \\
		R-D-F-50          & 36.29 & 25.43 & \underline{48.56} & \underline{69.31} & 2593 & \underline{70.87} \\
		\midrule
		\rowcolor{tblue} \textbf{TIDE-L}            & 36.64 & \underline{25.46} & 47.68 & 68.62 & 2809 & 70.06 \\
		\rowcolor{tblue} \textbf{TIDE-S}            & \textbf{38.36} & \textbf{26.40} & 45.52 & 66.15 & \textbf{2131} & 67.96 \\ \bottomrule
	\end{tabular*}
    \caption{Ablation study on feature extraction backbones. Abbreviations: R (ResNet), D (DCN), F (FPN).}
	\label{tab:backbone}
\end{table}

\subsubsection{Component Analysis of AGCIoU}

We evaluate AGCIoU by progressively ablating its core cues: Centroid Distance, Structural Consistency, and the Adaptive mechanism (Table~\ref{tab:iou}). Distance establishes a strong baseline, and structural cues yield further gains, theoretically aligning with the evolution of advanced metrics (e.g., CIoU) and inherently validating our geometric foundations without redundant comparisons. Finally, the Adaptive component acts as a state-aware safeguard. Although its overall performance gains are modest and introduce minor precision noise (a slight IDF1 drop), it effectively prevents geometric penalties from corrupting confident bounding-box overlaps. By selectively activating only during severe occlusions, it maximizes overall trajectory completeness (achieving the highest HOTA) without introducing complex learnable overhead.

\begin{table}[t]
	\centering
	\small
	\setlength{\tabcolsep}{4pt}
	
	\begin{tabular*}{\linewidth}{@{\extracolsep{\fill}} ccc|cccc}
		\toprule
		\textbf{Dist} & \textbf{S/R} & \textbf{Adapt} & \textbf{HOTA$\uparrow$} & \textbf{IDF1$\uparrow$} & \textbf{MOTA$\uparrow$} & \textbf{MOTP$\uparrow$} \\ \midrule
		\ding{51}     &            &         &    29.64     &    \underline{39.04}    &    \underline{54.42}    &    \underline{65.71}    \\
		\ding{51}     &      \ding{51}       &          &    \underline{29.94}     &    \textbf{39.79}    &    \textbf{54.43}    &    65.69    \\
		\rowcolor{tblue} \ding{51}     &      \ding{51}       &     \ding{51}     &    \textbf{29.98}     &    39.01    &    54.74    &    \textbf{65.82}    \\ \bottomrule
	\end{tabular*}
	
	\vspace{0.15cm}
	
	\begin{tabular*}{\linewidth}{@{\extracolsep{\fill}} ccc|ccccc}
		\toprule
		\textbf{Dist} & \textbf{S/R} & \textbf{Adapt} & \textbf{IDP$\uparrow$} & \textbf{IDR$\uparrow$} & \textbf{DetRe$\uparrow$} & \textbf{DetPr$\uparrow$} & \textbf{IDs$\downarrow$} \\ \midrule
		\ding{51}     &            &         &  \underline{48.01}    &   32.90    &    42.59     &    \underline{62.15}     &     \underline{862}      \\
		\ding{51}     &      \ding{51}       &          &  \textbf{48.90}    &   \textbf{33.53}    &    \underline{42.60}     &    62.12     &     \textbf{846}      \\
		\rowcolor{tblue} \ding{51}     &      \ding{51}       &     \ding{51}     &  47.87    &   \underline{32.92}    &    \textbf{42.86}     &    \textbf{62.32}     &     908      \\ \bottomrule
	\end{tabular*}
    \caption{Ablation study on the components of the IoU matching strategy. Abbreviations: Dist: Centroid proximity; S/R: Scale/Ratio (Structural consistency); Adapt: Adaptive weighting.}
	\label{tab:iou}
\end{table}

\subsubsection{Refinement Module}

We evaluated the contribution of the specialized C3TR and CSConv modules to analyze their impact on the accuracy-efficiency trade-off. Table~\ref{tab:Refinement Module} shows that while each module adds minimal computational cost, they both provide incremental performance gains. Specifically, their joint deployment yields a synergistic effect that elevates the IDF1 score to 29.55, demonstrating that local attention and feature redundancy reduction are highly complementary for resolving visual ambiguities among homogeneous targets. The full configuration achieves the best results, highlighting its effectiveness in enhancing tracking accuracy within strict lightweight constraints.

\begin{table}[t]
	\centering
	\small
	\setlength{\tabcolsep}{2.2pt}
	
	\begin{tabular*}{\linewidth}{@{\extracolsep{\fill}} cc|cc|ccc}
		\toprule
		\textbf{C3TR} & \textbf{CSConv} & \textbf{Params$\downarrow$} & \textbf{FLOPs$\downarrow$} & \textbf{HOTA$\uparrow$} & \textbf{IDF1$\uparrow$} & \textbf{MOTA$\uparrow$} \\ \midrule
		\ding{51} &  & \textbf{5.32M} & \textbf{20.17G} & 26.06 & 28.17 & 59.05 \\
		& \ding{51} & \underline{5.53M} & \underline{20.30G} & \underline{26.51} & \underline{28.33} & \underline{60.77} \\
		\rowcolor{tblue} \ding{51} & \ding{51} & 5.79M & 20.47G & \textbf{27.50} & \textbf{29.55} & \textbf{61.02} \\ \bottomrule
	\end{tabular*}
	
	\vspace{0.15cm}
	
	\begin{tabular*}{\linewidth}{@{\extracolsep{\fill}} cc|cccccc}
		\toprule
		\textbf{C3TR} & \textbf{CSConv} & \textbf{IDP$\uparrow$} & \textbf{IDR$\uparrow$} & \textbf{DetRe$\uparrow$} & \textbf{DetPr$\uparrow$} & \textbf{IDs$\downarrow$} & \textbf{MOTP$\uparrow$} \\ \midrule
		\ding{51} &  & 34.07 & 23.54 & 45.90 & \underline{68.39} & \underline{2789} & \underline{69.89} \\
		& \ding{51} & \underline{34.56} & \underline{24.00} & \underline{47.35} & 68.17 & \textbf{2746} & 69.81 \\
		\rowcolor{tblue} \ding{51} & \ding{51} & \textbf{36.04} & \textbf{25.04} & \textbf{47.76} & \textbf{68.75} & 2848 & \textbf{70.15} \\ \bottomrule
	\end{tabular*}
    \caption{Ablation study on refinement modules.}
	\label{tab:Refinement Module}
\end{table}

\subsubsection{Density Map Regression}

To evaluate the density map regression head, we compared the proposed model against a variant utilizing a standard head without density regression (Table~\ref{tab:density_head}). Integrating this module boosts HOTA from 26.76 to 28.43 and IDF1 from 31.10 to 36.29, alongside an expected decrease in MOTA and DetRe. This divergence reflects a deliberate architectural trade-off. Without the density constraint, the detector is overly sensitive in dense clusters, generating erratic bounding boxes that artificially inflate DetRe and MOTA but severely compromise tracking stability, leading to 2,718 ID switches. By constraining detection outputs against the global spatial distribution, the density head acts as a spatial regularizer to suppress unstable false positives. Consequently, the associator receives highly precise detections, which reduces ID switches by 79\% (to 574). These results confirm that trading erratic raw recall for stringent spatial precision is essential for identity maintenance in dense tracking scenarios.

\begin{table}[t]
	\centering
	\small
	\setlength{\tabcolsep}{4pt}
	
	\begin{tabular*}{\linewidth}{@{\extracolsep{\fill}} c|cccc}
		\toprule
		\textbf{Density Head} & \textbf{HOTA$\uparrow$} & \textbf{IDF1$\uparrow$} & \textbf{MOTA$\uparrow$} & \textbf{MOTP$\uparrow$} \\ \midrule
		                      & 26.76          & 31.10          & \textbf{71.15}          & 63.64          \\
		\rowcolor{tblue} \ding{51} & \textbf{28.43} & \textbf{36.29} & 47.84 & \textbf{67.17} \\ \bottomrule
	\end{tabular*}
	
	\vspace{0.15cm}
	
	\begin{tabular*}{\linewidth}{@{\extracolsep{\fill}} c|ccccc}
		\toprule
		\textbf{Density Head} & \textbf{IDP$\uparrow$} & \textbf{IDR$\uparrow$} & \textbf{DetRe$\uparrow$} & \textbf{DetPr$\uparrow$} & \textbf{IDs$\downarrow$} \\ \midrule
		                      & 33.93         & \textbf{28.71}         & \textbf{51.94}          & 61.40          & 2718          \\
		\rowcolor{tblue} \ding{51} & \textbf{49.44} & 28.67 & 37.34 & \textbf{64.41} & \textbf{574} \\ \bottomrule
	\end{tabular*}
	\caption{Ablation study on the effectiveness of the density map regression head.}
	\label{tab:density_head}
\end{table}

\subsection{Generalization on MFT25 Benchmark}
\label{sec:Generalization}

To evaluate cross-domain generalization, we validate our framework on the large-scale MFT25 benchmark (Table~\ref{tab:tracking-comparison}). Heavy SDE paradigms like SU-T naturally achieve higher absolute accuracy (34.07 HOTA). However, their reliance on massive 99.00M parameter backbones and computationally expensive Re-ID networks (793.21G FLOPs) strictly prohibits application on resource-constrained edge platforms. Rather than competing under unlimited computing assumptions, TIDE prioritizes real-world feasibility. The lightweight L-branch establishes a crucial edge-deployment baseline: trading a 40.9\% HOTA reduction compared to SU-T to completely eliminate 97.4\% of the required FLOPs and 94.1\% of the parameters, translating an impossible edge-tracking task into a deployable reality.

Furthermore, analyzing this trade-off exclusively through the L-branch obscures the inherent scalability of the proposed architecture. By exploring the Pareto frontier of computational efficiency and tracking robustness, the scalable S-branch demonstrates how a moderate structural expansion yields substantial precision recoveries. Elevating the HOTA to 27.13, TIDE-S significantly outperforms classic JDE baselines such as FairMOT (22.22) and CMFTNet (22.43). This parallel evaluation confirms that our dual-branch elasticity effectively navigates the accuracy-deployability trade-off, proving that strict geometric association can reliably substitute heavy appearance features under tight computational constraints.

\begin{table}[t]
	\centering
	\small
	\setlength{\tabcolsep}{3pt} 
	\begin{tabular*}{\linewidth}{@{\extracolsep{\fill}} l|cc|ccc}
		\toprule
		\textbf{Methods} & \textbf{Params$\downarrow$} & \textbf{FLOPs$\downarrow$} & \textbf{HOTA$\uparrow$} & \textbf{IDF1$\uparrow$} & \textbf{MOTA$\uparrow$} \\ \midrule
		FairMOT          & \underline{16.55M} & \underline{72.93G}  & 22.226 & 26.867 & 47.509 \\
		CMFTNet          & 45.08M & 137.77G & 22.432 & 27.659 & 46.365 \\
		OC-SORT          & 99.00M & 793.21G & 25.017 & 34.620 & 46.706 \\
		TFMFT            & 39.93M & 215.27G & 25.440 & 33.950 & 49.725 \\
		TransCenter      & 30.66M & 133.09G & 27.896 & 30.278 & 68.693 \\
		SORT             & 99.00M & 793.21G & 29.063 & 34.119 & 69.038 \\
		TrackFormer      & 42.95M & 143.43G & 30.361 & 35.285 & \textbf{74.609} \\
		ByteTrack        & 99.00M & 793.21G & 31.758 & 40.355 & \underline{69.586} \\
		HybridSORT       & 99.00M & 793.21G & \underline{32.705} & \underline{41.727} & 69.167 \\ \midrule
		SU-T             & 99.00M & 793.21G & \textbf{34.067} & \textbf{44.643} & 68.958 \\ \midrule
		\rowcolor{tblue} \textbf{TIDE-L (Ours)}  & \textbf{5.79M}  & \textbf{20.47G} & 20.135 & 23.654 & 31.996 \\ 
		\rowcolor{tblue} \multicolumn{1}{r|}{\textit{$\Delta$ vs. SU-T}} & \textcolor{upgreen}{\textit{\textbf{-94.1\%}}} & \textcolor{upgreen}{\textit{\textbf{-97.4\%}}} & \textcolor{downred}{\textit{\textbf{-40.9\%}}} & \textcolor{downred}{\textit{\textbf{-47.0\%}}} & \textcolor{downred}{\textit{\textbf{-53.6\%}}} \\
		\rowcolor{tblue} \textbf{TIDE-S (Ours)}  & 32.59M & 90.13G  & 27.137 & 33.196 & 64.694 \\ 
		\rowcolor{tblue} \multicolumn{1}{r|}{\textit{$\Delta$ vs. SU-T}} & \textcolor{upgreen}{\textit{\textbf{-67.1\%}}} & \textcolor{upgreen}{\textit{\textbf{-88.6\%}}} & \textcolor{downred}{\textit{\textbf{-20.3\%}}} & \textcolor{downred}{\textit{\textbf{-25.6\%}}} & \textcolor{downred}{\textit{\textbf{-6.2\%}}} \\ \bottomrule
	\end{tabular*}
	\caption{Generalization evaluation and comparison of MFT and MOT methods on the MFT25 benchmark.}
	\label{tab:tracking-comparison}
\end{table}

\subsection{Qualitative Results and Visualization}

\begin{figure}[t]
	\centering
	\includegraphics[width=\linewidth]{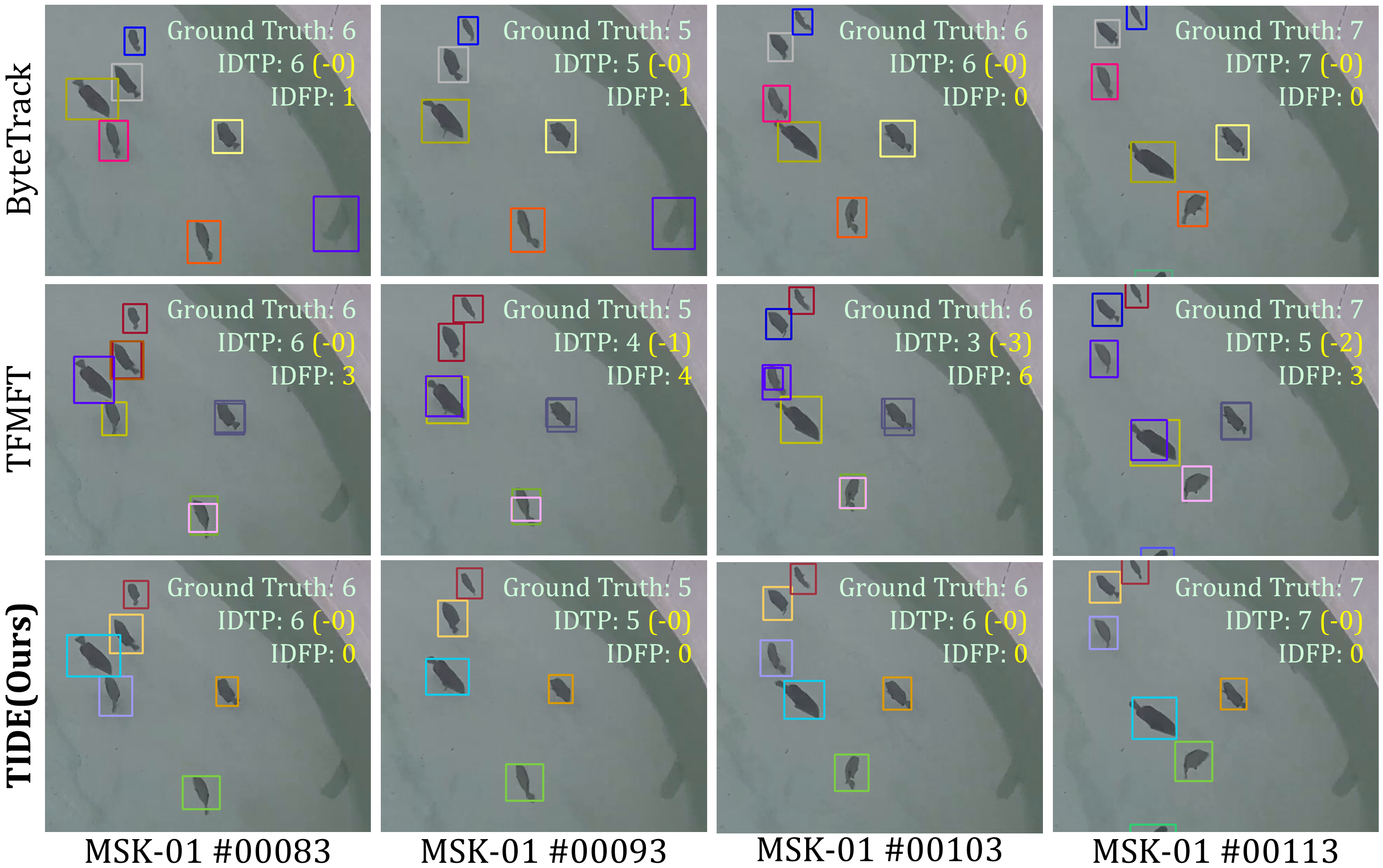}
	\caption{Tracking visualization in a recirculating aquaculture system. Prominent dynamic shadows often lead to repeated detections and misidentifications in baseline models. However, TIDE consistently maintains stable tracking without missed or incorrect detections. Best viewed in color.}
	\label{fig:MSK_Vis}
\end{figure}

To qualitatively assess the robustness of our framework, we visualized tracking results from challenging scenarios within the MFT-Edge dataset. Fig.~\ref{fig:MSK_Vis} illustrates a representative edge-case featuring prominent dynamic shadow interference in a recirculating aquaculture system. 

While baseline trackers like ByteTrack and TFMFT suffer from severe false positives and tracking fragmentation due to detector over-sensitivity to illumination changes, TIDE consistently delivers accurate localization. This stability is largely attributed to our shared multi-task head, which mutually constrains detection and density regression, allowing TIDE to infer locations accurately even when visual features are corrupted by shadows. 

Beyond this specific scenario, our broader qualitative analysis confirms that TIDE excels in mitigating environmental noise, mirror reflections, and non-rigid deformations through the AGCIoU metric. The primary qualitative limitation observed is that extreme and prolonged occlusions in highly dense clusters can still induce occasional ID switches, representing a common bottleneck for linear motion prediction models.
\section{Conclusion}

In this paper, we proposed TIDE, a highly efficient JDE framework featuring a scalable dual-branch architecture and a minimalist AGCIoU metric for high-density multiple fish tracking. To bridge the gap between heavy academic benchmarks and resource-constrained industrial applications, TIDE effectively resolves computational bottlenecks while maintaining robust ID consistency without heavy appearance models. Extensive evaluations demonstrate an exceptional accuracy-efficiency balance: the lightweight L-branch achieves a competitive 28.43 HOTA and reduces ID switches by 69.8\%, operating on merely 5.79M parameters to yield a 38.7-fold reduction in FLOPs compared to standard heavy trackers. Concurrently, the scalable S-branch further elevates precision to 29.98 HOTA and 39.01 IDF1 for demanding server-side applications. While our results confirm strong cross-domain generalization, future work will explore efficient non-linear prediction models to address extreme morphological variations and complex motion drift across diverse aquatic environments.


\setcounter{figure}{0}
\setcounter{table}{0}
\setcounter{equation}{0}
\renewcommand{\thefigure}{S\arabic{figure}}
\renewcommand{\thetable}{S\arabic{table}}
\renewcommand{\theequation}{S\arabic{equation}}
\renewcommand{\thesection}{\Alph{section}}

\section*{Appendix A.~Detailed Pipeline Architecture}

\begin{figure*}[t]
	\centering
	\includegraphics[width=\linewidth]{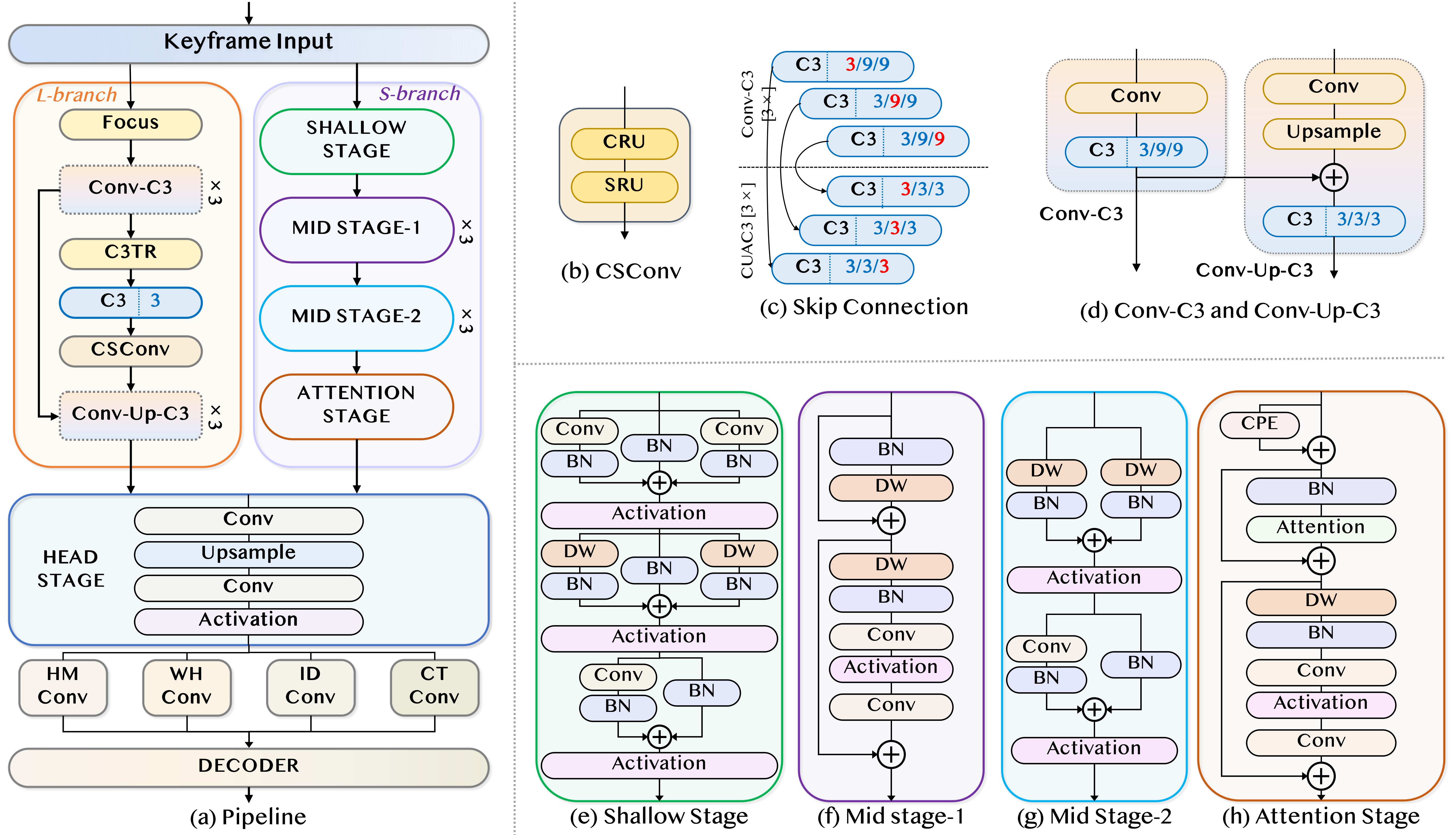}
	\caption{Detailed architecture of the dual-branch feature extraction pipeline. (a) gives an overview. Key modules like (b) CSConv and (d) Conv-C3, Conv-Up-C3 are detailed. (c) shows skip connections, where the number sequences denote the stacking depth of C3 modules, with highlighted digits indicating the exact matching pairs. The Scalable S-branch (e-h) is a deep, hybrid CNN-Transformer architecture, while the Lightweight L-branch is a highly efficient CNN-based design.}
	\label{fig:S_Pipeline}
\end{figure*}

Due to space constraints in the main manuscript, the detailed layer-wise operations of our dual-branch feature extraction pipeline are presented in this section. As shown in Fig.~\ref{fig:S_Pipeline}, the pipeline is engineered with true computational elasticity to accommodate varying deployment scenarios.

\subsection*{A.1. Scalable branch (S-branch)}
The S-branch is engineered to achieve maximum tracking accuracy on high-capacity hardware by capturing both rich local features and long-range global dependencies. It employs a hybrid CNN and Transformer architecture. 
In the \textit{shallow stage}, efficient low-level features like edges and textures are extracted using standard and depthwise convolutions. The \textit{mid stage} is repeated three times to progressively learn increasingly abstract representations. Here, we employ a large $7\times7$ convolution to expand the receptive field with minimal parameter increase, capturing the entire fish body even at a distance. The \textit{attention stage} leverages the strength of Vision Transformers to model long-range dependencies, vital for distinguishing between similar-looking fish in crowded and occluded scenes.

\subsection*{A.2. Lightweight branch (L-branch)}
The L-branch maximizes computational efficiency for real-time edge deployment on resource-constrained hardware. It begins with a Focus operation, compressing the spatial resolution of high-definition underwater frames into channel dimensions without discarding critical low-level textural details. The primary spatial feature extraction is driven by a sequence of Conv-C3 groups, alternately applying spatial downsampling and residual learning.
To achieve robust multi-scale feature fusion without the heavy computational overhead of a traditional feature pyramid network, we introduce the Conv-Up-C3 module. It integrates $1\times1$ convolution layers for channel adjustment, nearest-neighbor upsampling for spatial enlargement, and skip connections via concatenation to fuse deep semantic features with shallow maps. Finally, the C3TR module (attention-based Transformer block) and the CSConv module (reconstructing spatial and channel features) are incorporated to enhance tracking accuracy while strictly preserving efficiency bounds.

\subsection*{A.3. Shared Multi-task Head}
Both branches feed into a shared \textit{head stage}. This stage uses standard convolutions and upsampling to prepare features for four parallel sub-tasks: heatmap generation, bounding box prediction, appearance feature extraction, and counting. A unified decoder then extracts the final outputs, including object centroids, dimensions, offsets, and total counts.

\section*{Appendix B.~Detailed Mathematical Formulations}

This section provides the comprehensive mathematical derivations and formulations for the proposed AGCIoU metric and the multi-task training objective function not fully covered in the main text.

\subsection*{B.1. AGCIoU Formulation}
The AGCIoU calculation begins by computing the fundamental dimensions of the bounding boxes:
\begin{equation}
	\footnotesize
	\begin{aligned}
		w_i = x_{i2} - x_{i1}, \quad h_i = y_{i2} - y_{i1}, \quad A_i = w_i\,h_i\\
	\end{aligned}
	\label{eq:S_bbox_size}
\end{equation}
where $(x_{i1}, y_{i1})$ and $(x_{i2}, y_{i2})$ represent the coordinates of the top-left and bottom-right corners of bounding box $i$. Traditional IoU is measured as:
\begin{equation}
	\footnotesize
	\begin{aligned}
		\mathrm{IoU}
		= \frac{|B_1 \cap B_2|}{|B_1 \cup B_2| + \varepsilon}
	\end{aligned}
	\label{eq:S_tradition_iou}
\end{equation}

To establish the centroid proximity, we calculate the Euclidean distance between the bounding box centroids and normalize it by the average geometric scale to ensure scale invariance:
\begin{equation}
	\footnotesize
	\begin{aligned}
		(x_{c,i},\,y_{c,i}) = \Bigl(\tfrac{x_{i1}+x_{i2}}{2},\,\tfrac{y_{i1}+y_{i2}}{2}\Bigr), 
		\\ \quad d_c = \sqrt{(x_{c,2}-x_{c,1})^2 + (y_{c,2}-y_{c,1})^2}
	\end{aligned}
	\label{eq:S_centroid_distance}
\end{equation}
\begin{equation}
	\footnotesize
	\begin{aligned}
		s = \tfrac{\sqrt{A_1} + \sqrt{A_2}}{2}
	\end{aligned}
	\label{eq:S_normalized_distance}
\end{equation}
Based on these foundational measurements, the normalized center distance ($\delta$), aspect ratio alignment factor ($\alpha$), and area size alignment factor ($s_r$) are calculated as detailed in the main manuscript to compute the final distance and shape penalties.

\subsection*{B.2. Multi-task Loss Functions}
The network is jointly optimized for target localization, bounding box regression, identity classification, and scene counting. The heatmap loss $L_{hm}$ supervises target centroids using Focal Loss:
\begin{equation}
	\footnotesize
	\begin{aligned}
		L_{hm} =-\frac{1}{n}\sum_{xy}
		\begin{cases}
			(1-\hat{M}_{xy})^{\alpha}\log(\hat{M}_{xy}), & M_{xy}=1;\\ 	
			\begin{array}{l}
			(1-M_{xy})^{\beta}(\hat{M}_{xy})^{\alpha} \\
			\times \log(1-\hat{M}_{xy}),
			\end{array} & \text{otherwise}.
		\end{cases}
	\end{aligned}
	\label{eq:S_heatmap}
\end{equation}

The bounding box size and centroid offset are regressed using L1 loss:
\begin{equation}
	\footnotesize
	\begin{aligned}
		L_{bs+os}=\sum_{i=1}^{n}\left|s^{i}-\hat{s}^{i}\right|+\lambda\left|o^{i}-\hat{o}^{i}\right|
	\end{aligned}
	\label{eq:S_box_off}
\end{equation}

The ID appearance classification is supervised using a cross-entropy loss:
\begin{equation}
	\footnotesize
	\begin{aligned}
		L_{id}=-\sum_{i=1}^n\sum_{k=1}^K\mathcal{L}^i(k)\mathrm{log}(\mathrm{p}(k))
	\end{aligned}
	\label{eq:S_id}
\end{equation}

The density map regression and auxiliary mutually restraining constraints ($L_{ct}$, $L_{\tilde{det}}$, and $L_{\tilde{ct}}$) are formulated and fused dynamically via the uncertainty-driven weighting strategy specified in the main text. This adaptive mechanism ensures balanced gradient propagation across all branches without requiring manual scaling heuristics.

\section*{Appendix C.~Detailed Trajectory Association Flowcharts}

The main text introduces our cascaded two-stage matching strategy. This strategy systematically decouples appearance-based matching from spatial-geometric matching to enhance trajectory prediction accuracy while minimizing computational overhead. We provide detailed visual flowcharts for both association stages.

\subsection*{C.1. First Stage: Appearance-based Matching}
\begin{figure*}
	\centering
	\includegraphics[width=\linewidth]{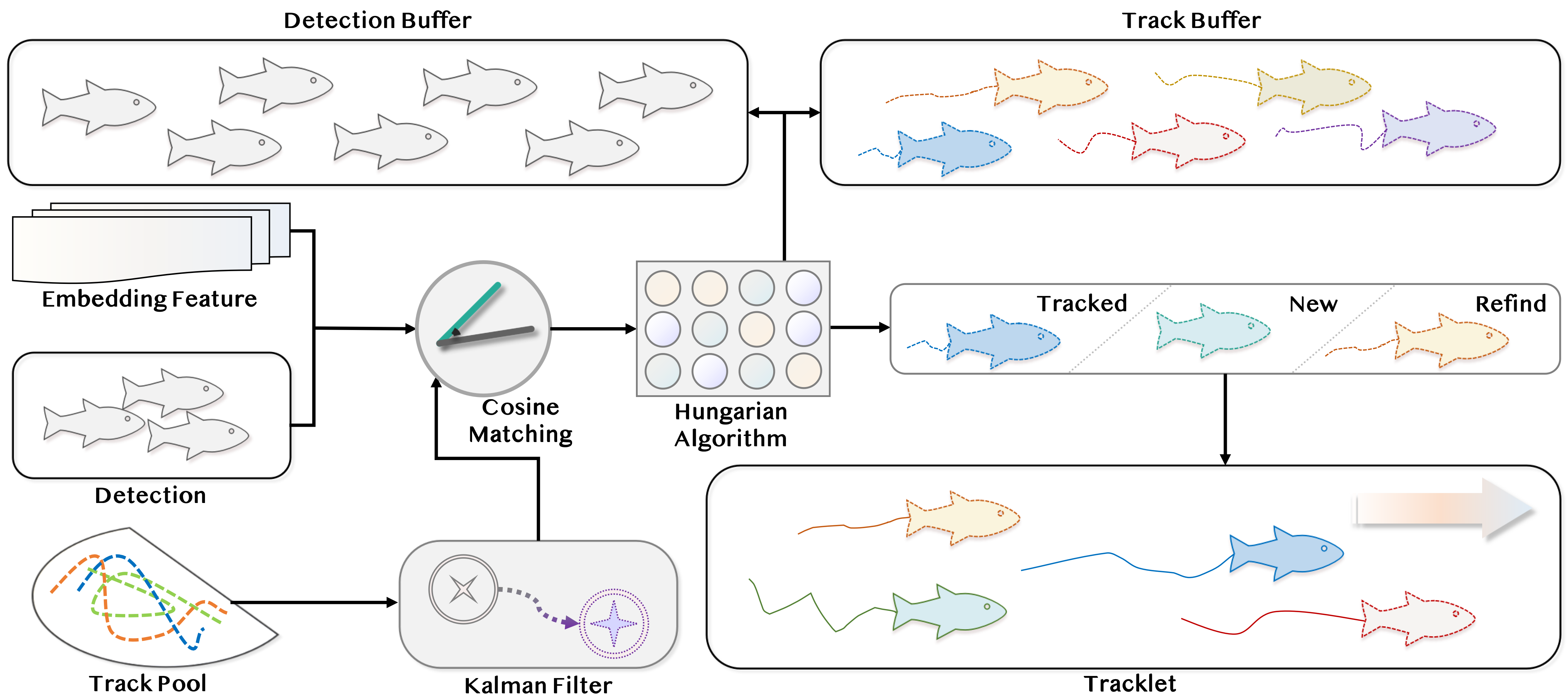}
	\caption{Visualization of the first stage (Appearance Matching). Detections and existing tracks are fed into a Kalman Filter for motion prediction. An initial association is performed by matching appearance feature embeddings using the Hungarian algorithm.}
	\label{fig:S_Asso1}
\end{figure*}

Fig.~\ref{fig:S_Asso1} illustrates the first stage, which performs an initial, high-confidence matching utilizing the embedded appearance features generated by the multi-task head. Data flows from both the \textit{Detection Buffer} and \textit{Track Buffer}, while historical trajectories from the \textit{Track Pool} simultaneously undergo \textit{Kalman Filter} prediction to temporally align the matching space. A cosine distance module evaluates the semantic similarity between the newly extracted appearance embeddings and the existing tracklet features, which the \textit{Hungarian Algorithm} then processes to resolve optimal global assignments. By isolating and resolving these unambiguous assignments early, the framework significantly reduces the computational burden on the subsequent geometric matching phase. Successfully matched instances bypass the second stage entirely and are directly updated within the final \textit{Tracklet Pool} as continuously tracked or successfully refound targets.

\begin{figure*}
	\centering
	\includegraphics[width=\linewidth]{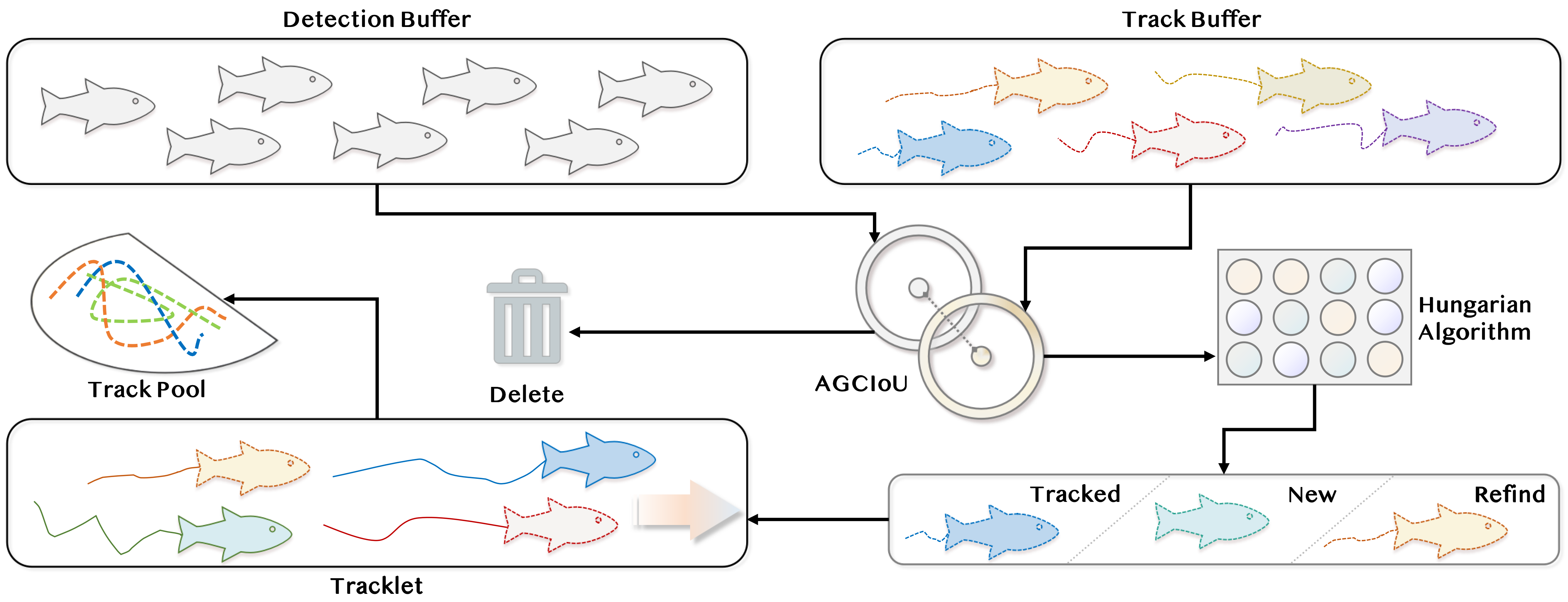}
	\caption{Visualization of the second stage (AGCIoU Matching). Unmatched detections and tracks are processed using the minimalist AGCIoU metric. A two-tier strategy first matches high-confidence detections, followed by low-confidence ones, before initializing new tracks.}
	\label{fig:S_Asso2}
\end{figure*}

\subsection*{C.2. Second Stage: Geometric Matching with AGCIoU}
Fig.~\ref{fig:S_Asso2} details the second stage, acting as a robust safety net for targets that remain unmatched due to severe physical adhesion or rapid morphological changes. To prevent error accumulation, it explicitly abandons appearance embeddings and performs robust geometric matching using our Adaptive Geometric Correspondence IoU (AGCIoU) metric. The flowchart highlights our rigorous two-tier recovery strategy: it first associates remaining unmatched tracks with high-confidence detections to stabilize visually ambiguous targets, followed by a secondary matching pass utilizing low-confidence detections to salvage heavily occluded fish. Finally, the associator executes strict trajectory lifecycle management by initializing unmatched high-confidence detections as new tracks and permanently purging unrecovered lost tracks, represented by the \textit{Delete} module in the visual pipeline. This closed-loop mechanism ensures that the active track pool remains highly efficient and resilient against accumulated noise over extended observations.

\section*{Appendix D.~Sensitivity Analysis}

We performed a comprehensive sensitivity analysis on the key matching and tracking thresholds utilized during the association phase. This analysis ensures the optimal performance and robustness of our framework. 

\begin{figure}[htbp]
	\centering
	\includegraphics[width=0.8\linewidth]{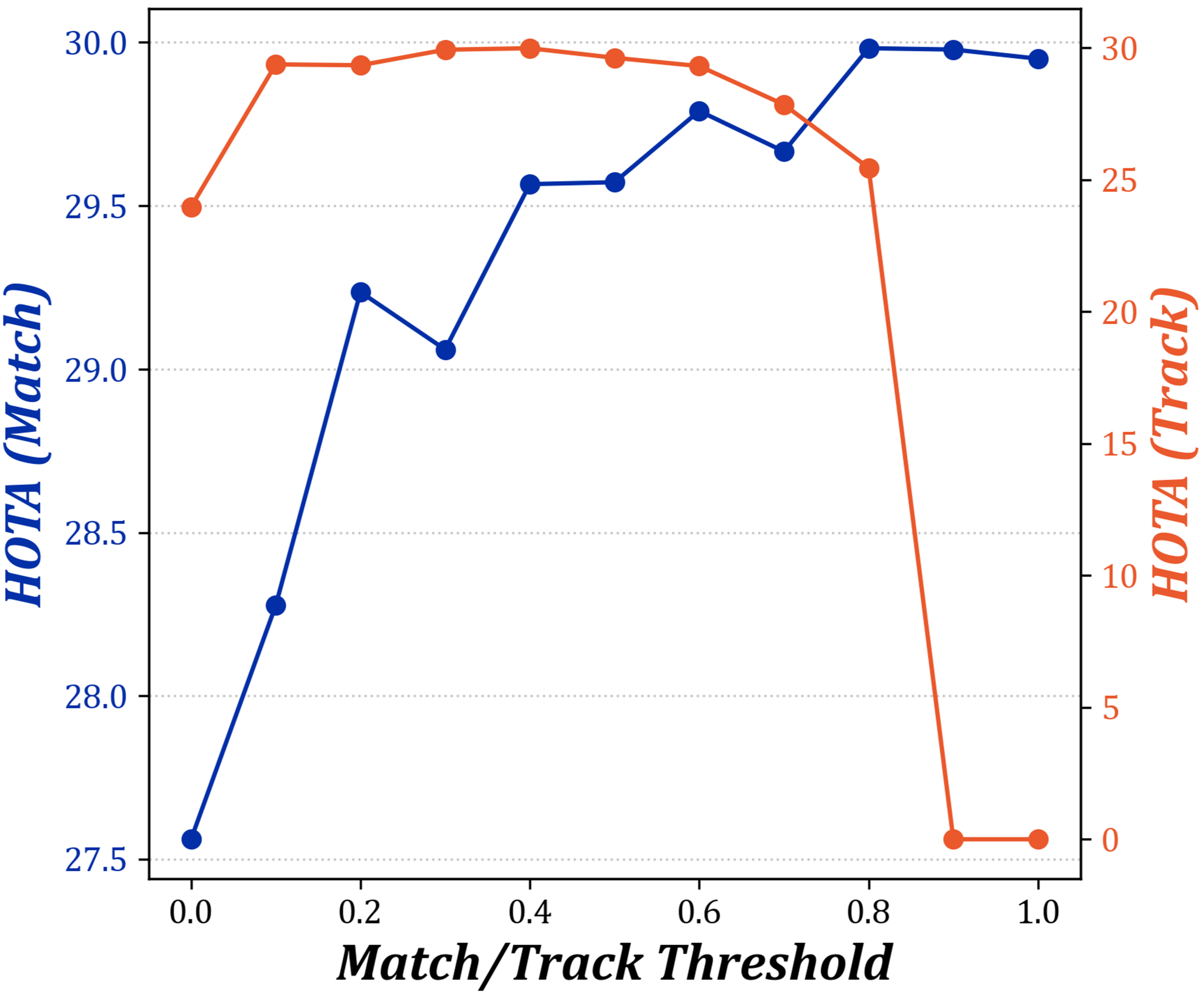}
	\caption{Impact of matching and tracking thresholds on tracking sensitivity performance.}
	\label{fig:S_Threshold}
\end{figure}

As illustrated in Fig.~\ref{fig:S_Threshold}, tracking performance (measured by HOTA) peaks when the matching threshold is between 0.8 and 1.0. This confirms that the AGCIoU metric is highly reliable under strict geometric correspondence. Concurrently, tracking remains stable across tracking thresholds ranging from 0.1 to 0.6. Based on these empirical findings, we set the default matching and tracking thresholds to 0.8 and 0.4, respectively, for all experiments reported in the main text. These optimal values demonstrate strong generalization across the MFT-Edge dataset. Nevertheless, they may be domain-specific. Applying the framework to radically different aquaculture environments or disparate aquatic species could require fine-tuning.

\section*{Appendix E.~Additional Qualitative Visualizations}

Due to strict space limitations in the main manuscript, we only presented visualization results for the recirculating aquaculture system sequence featuring prominent dynamic shadows. In this section, we provide extensive qualitative assessments across other challenging scenarios within the MFT-Edge dataset to further validate the robustness of the TIDE framework. We compare the actual number of fish (Ground Truth) with the correctly detected instances (IDTP) and false detections (IDFP).

\begin{figure}[htbp]
	\centering
	\includegraphics[width=\linewidth]{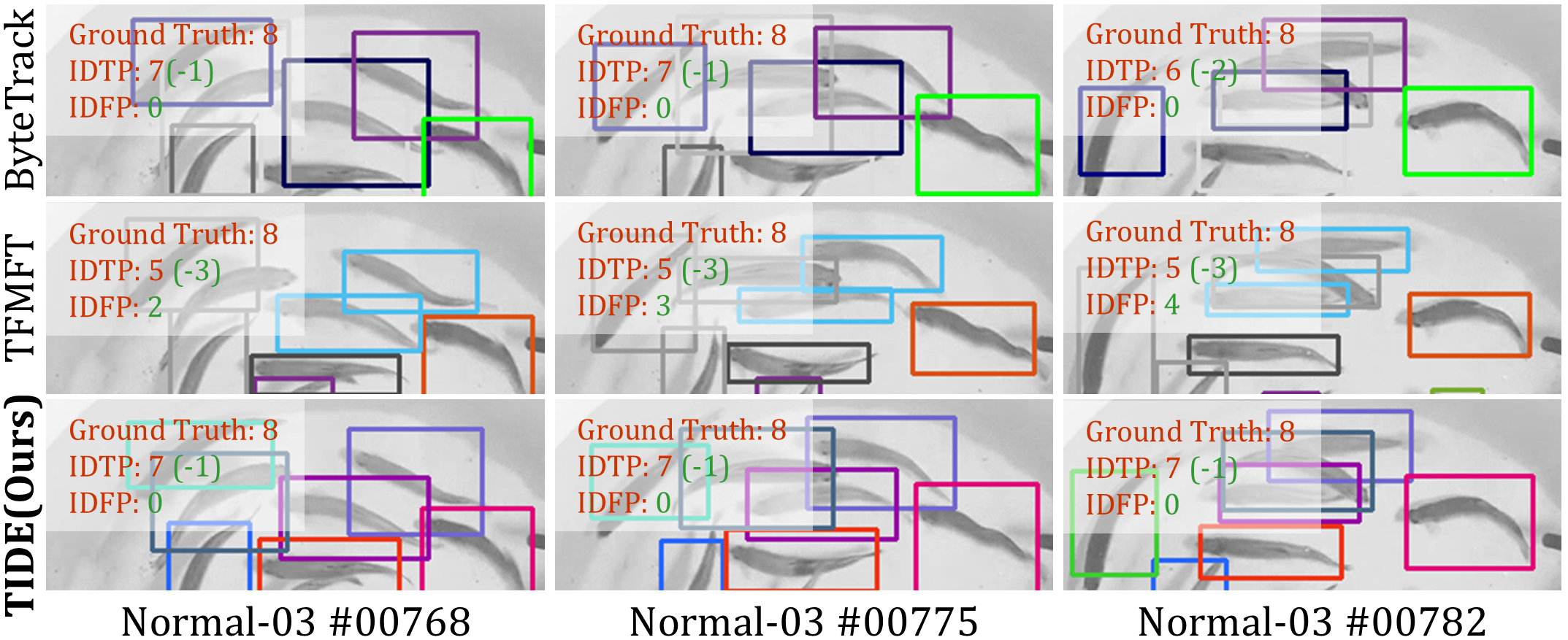}
	\caption{Tracking results of the Normal-03 clip. The fish are visibly clustered together and swimming rapidly. TIDE demonstrates zero false detections and robust identity maintenance against severe adhesion.}
	\label{fig:S_N3_Vis}
\end{figure}

\paragraph{Dense Adhesion and Rapid Circular Motion (Normal-03):}

Fig.~\ref{fig:S_N3_Vis} showcases crucian carp fry swimming rapidly in a circular motion. Although lighting conditions are optimal, the severe physical adhesion between targets poses a significant challenge. TIDE achieves zero false detections and misses only one target (at frame 775). In comparison, TFMFT exhibits a false detection rate ranging from 25\% to 50\%. It consistently shows three missed detections across the displayed keyframes.

\begin{figure}[htbp]
	\centering
	\includegraphics[width=\linewidth]{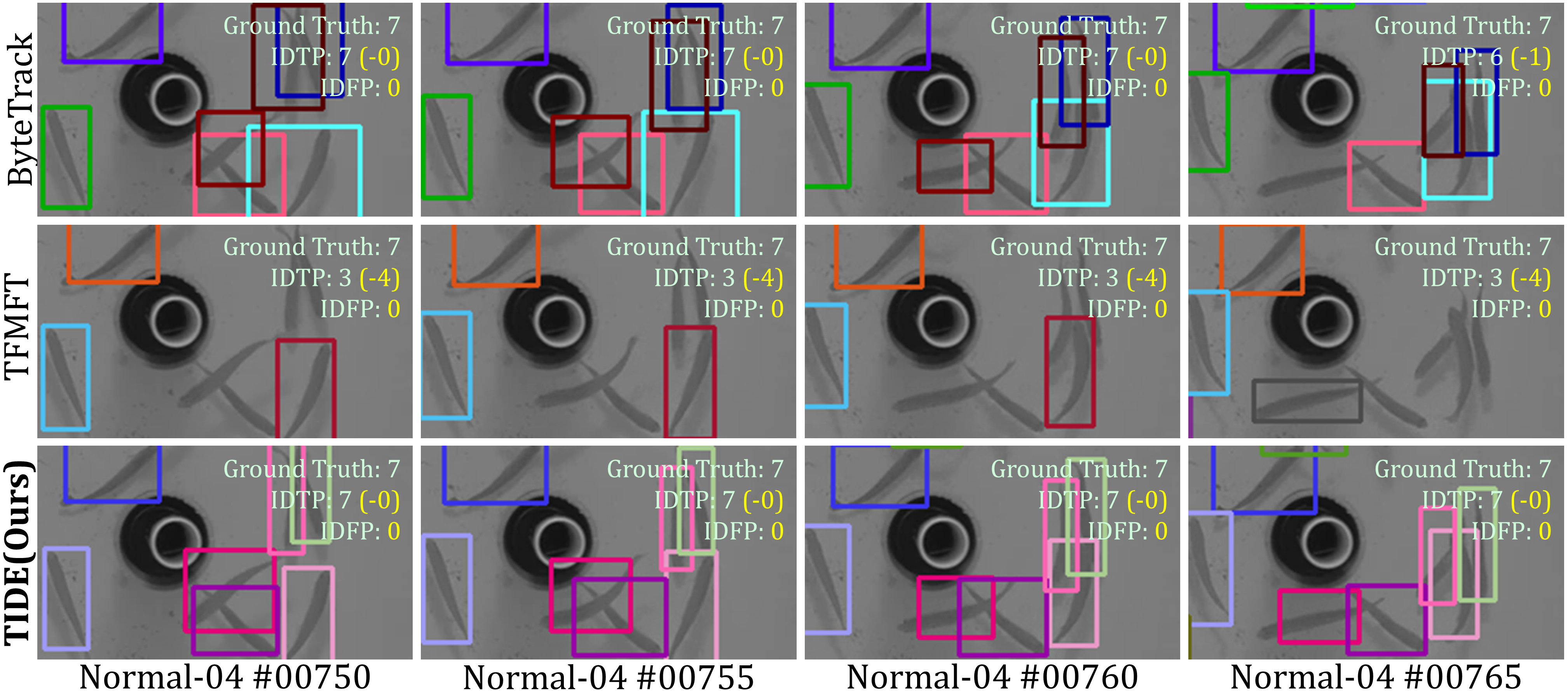}
	\caption{Performance on the Normal-04 clip. The fish exhibit irregular swimming patterns with significant changes in body shape. TIDE maintained stable tracking, highlighting its robustness against non-rigid deformations.}
	\label{fig:S_N4_Vis}
\end{figure}

\paragraph{Irregular Swimming and Morphological Deformation (Normal-04):}
\begin{figure*}[!t]
	\centering
	\includegraphics[width=\linewidth]{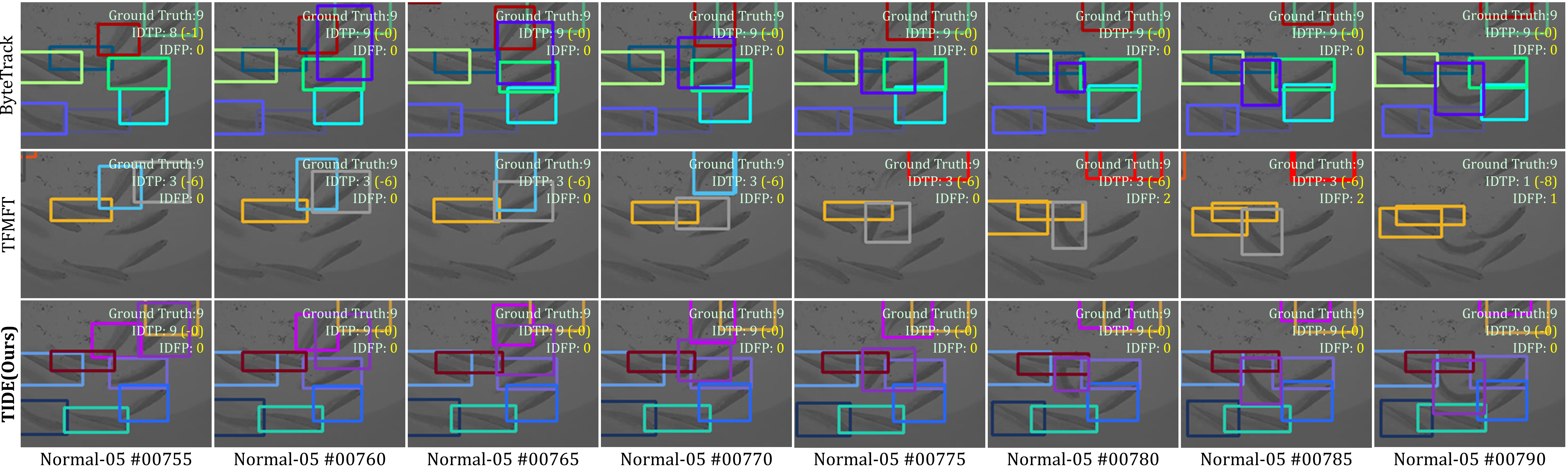}
	\caption{Tracking visualization on the Normal-05 clip. Despite the extremely low brightness and contrast, poor fish visibility, and interference from residual bait, TIDE maintained stable performance without any missed or incorrect detections. Different colored bounding boxes indicate distinct individual fish.}
	\label{fig:S_N5_Vis}
\end{figure*}

\begin{figure*}[!t]
	\centering
	\includegraphics[width=\linewidth]{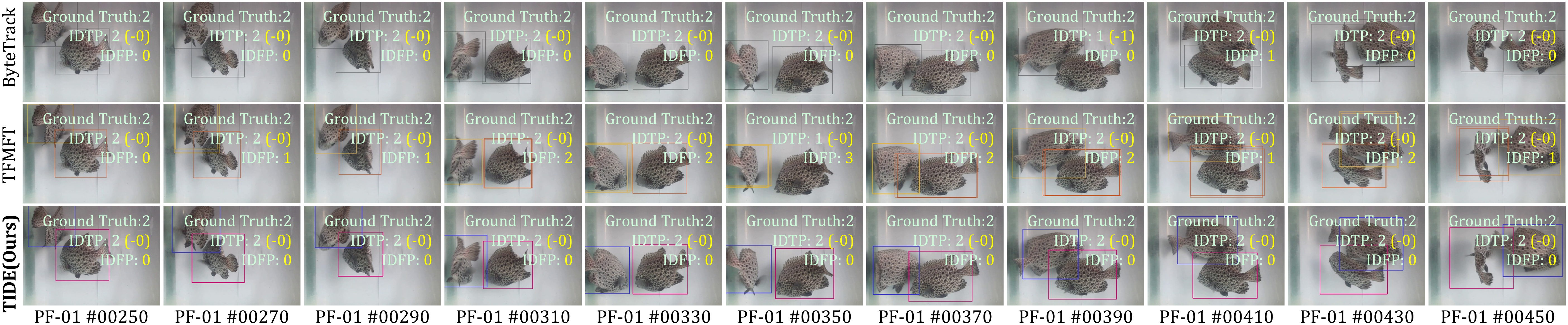}
	\caption{Illustration of results on the PF-01 video clip. Fish individuals are notably affected by mirror reflections, leading to significant occlusions and misalignments. TIDE maintains stable IDs without missed or false detections.}
	\label{fig:S_PF_Vis}
\end{figure*}

Fig.~\ref{fig:S_N4_Vis} illustrates performance when crucian carp fry swim irregularly, presenting significant body shape variations (e.g., C-shaped bending) under moderate lighting. ByteTrack performs commendably with only one missed detection. Conversely, TFMFT exhibits four irrecoverable missed detections starting from frame 750. TIDE achieves flawless tracking with zero missed or false detections. This underscores the efficacy of the AGCIoU metric in handling drastic morphological deformations.

\paragraph{Ultra-Low Light and Contrast (Normal-05):} 
Fig.~\ref{fig:S_N5_Vis} illustrates tracking performance in a severely low-brightness scene. The color contrast between the fish and the background is minimal, accompanied by significant interference from residual feed. Among the evaluated methods, TIDE exhibited zero false detections throughout the sequence. ByteTrack missed a detection at frame 755 before recovering. Conversely, TFMFT showed severe degradation, experiencing over six missed detections and repeated false positives starting from frame 780.

\paragraph{Severe Mirror Reflections (PF-01):}
Fig.~\ref{fig:S_PF_Vis} shows performance under conditions of severe optical interference, featuring spotted knifejaw in a transparent fish tank. The fish are notably affected by mirror reflections, leading to significant occlusions and visual misalignments. TIDE consistently maintains accurate ID tracking throughout the sequence. In contrast, TFMFT suffers from persistent duplicate and missed detections starting from frame 270. This failure is primarily driven by reflection-induced ambiguities.

{
	\small
	\bibliographystyle{ieeenat_fullname}
	\bibliography{ref}
}

\end{document}